\RequirePackage{fix-cm}
\documentclass{svjour3}                     
\smartqed  
\usepackage{pdfpages}
\usepackage{hyperref}
\usepackage{graphicx}
\usepackage{algorithmic}
\usepackage{mathtools}
\usepackage{multirow}
\usepackage{natbib}
\usepackage{marvosym}

\newcommand*{\affaddr}[1]{#1}
\newcommand*{\affmark}[1][*]{\textsuperscript{#1}}

\newcommand{\envelope}{(\raisebox{-.5pt}{\scalebox{1.45}{\Letter}}\kern-1.7pt)}

\journalname{Journal of Heuristics}

\begin{document}

\title{Roster Evaluation Based on Classifiers for the Nurse Rostering Problem
}


\author{Roman V\'{a}clav\'{i}k\affmark[1]        \and
        P\v{r}emysl \v{S}\r{u}cha\affmark[1] 		\and
        Zden\v{e}k Hanz\'{a}lek\affmark[1,2]
}


\institute{
R. V\'{a}clav\'{i}k \envelope \and P. \v{S}\r{u}cha \and Z. Hanz\'{a}lek
\at\email{\{vaclarom, suchap, hanzalek\}@fel.cvut.cz} \\ \\
\affaddr{\affmark[1]Department of Control Engineering, Faculty of Electrical Engineering, Czech Technical University, Karlovo n\'{a}m\v{e}st\'{i} 13, 121 35 Prague 2, Czech Republic}\\
\affaddr{\affmark[2]Czech Institute of Informatics, Robotics, and Cybernetics, Czech Technical University in Prague, Zikova street 1903/4, 166 36 Prague 6, Czech Republic}
}

\date{Received: date / Accepted: date}

\pagenumbering{gobble}
\includepdf[pages=1,fitpaper,noautoscale]{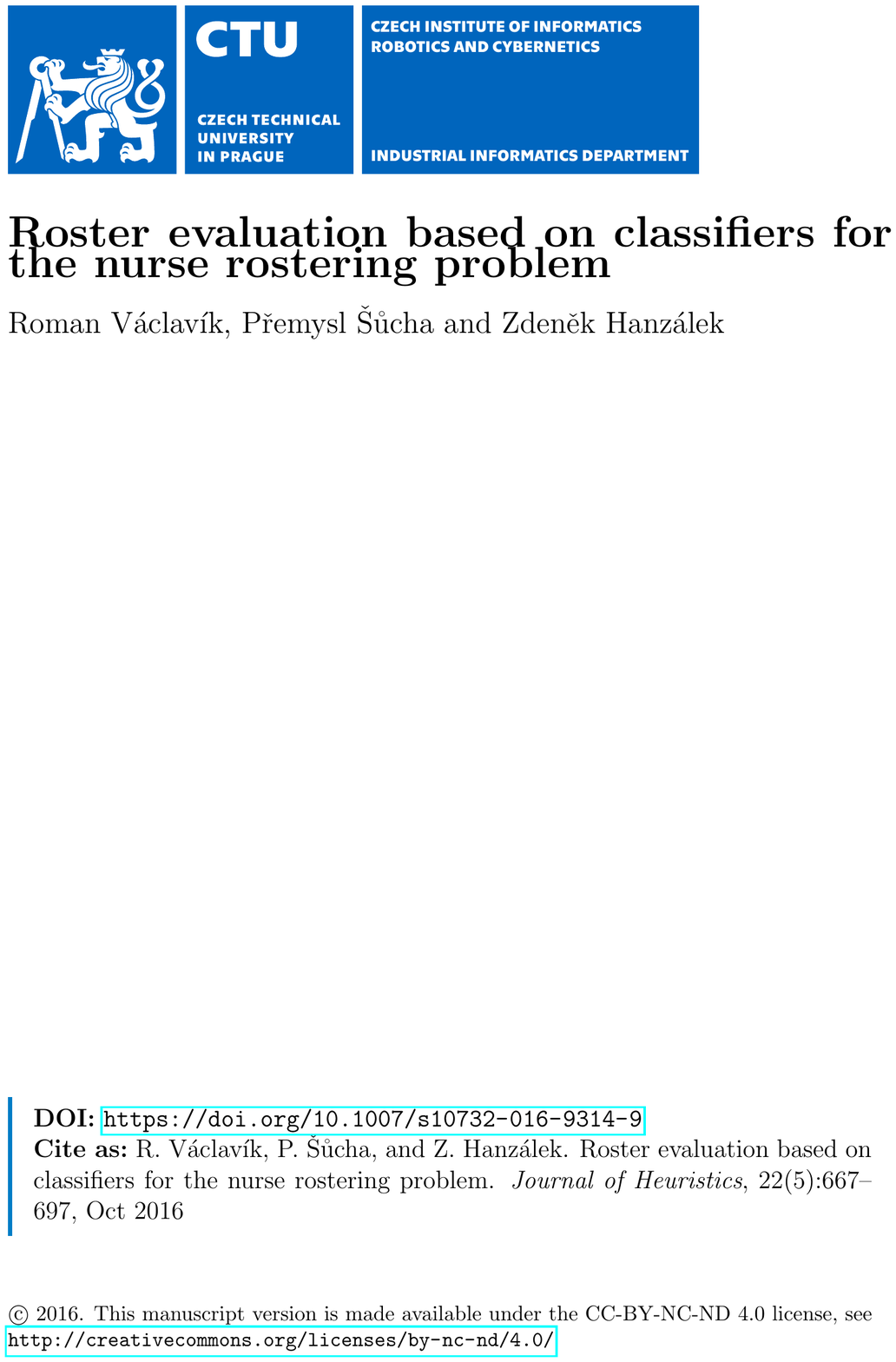}
\pagenumbering{arabic}
\maketitle

\begin{abstract}
The personnel scheduling problem is a well-known NP-hard combinatorial problem. Due to the complexity of this problem and the size of the real-world instances, it is not possible to use exact methods, and thus heuristics, meta-heuristics, or hyper-heuristics must be employed. The majority of heuristic approaches are based on iterative search, where the quality of intermediate solutions must be calculated. Unfortunately, this is computationally highly expensive because these problems have many constraints and some are very complex. In this study, we propose a machine learning technique as a tool to accelerate the evaluation phase in heuristic approaches. The solution is based on a simple classifier, which is able to determine whether the changed solution (more precisely, the changed part of the solution) is better than the original or not. This decision is made much faster than a standard cost-oriented evaluation process. However, the classification process cannot guarantee 100\% correctness. Therefore, our approach, which is illustrated using a tabu search algorithm in this study, includes a filtering mechanism, where the classifier rejects the majority of the potentially bad solutions and the remaining solutions are then evaluated in a standard manner. We also show how the boosting algorithms can improve the quality of the final solution compared with a simple classifier. We verified our proposed approach and premises, based on standard and real-world benchmark instances, to demonstrate the significant speedup obtained with comparable solution quality.
\keywords{neural network \and nurse rostering problem \and adaptive boosting \and pattern learning.}
\end{abstract}

\section{Introduction}
\label{sec:m}
Personnel scheduling, such as the nurse rostering problem (NRP), is a well-known combinatorial problem, which is known to be NP-hard \citep{Karp72,Aickelin00}. This problem involves the assignment of shifts to employees (nurses) and a solution to this problem is called a roster. Numerous studies have proposed exact and heuristic algorithms for solving the NRP \citep{Burke04}. Good rosters can save a significant amount of company money as well as improving employee satisfaction with their workload. In the present study, we focus on the use of a human-inspired method for determining the roster quality. The basic idea entails the rapid recognition of an obviously bad roster structure, which is a relatively easy task for an experienced human scheduler. The main goal is to acquire some knowledge from the previous runs of a rostering algorithm to accelerate the execution of the algorithm, where a significant portion of the time is spent evaluating the potential solutions.

\begin{figure}[ht]
\centering
\includegraphics[width=10cm]{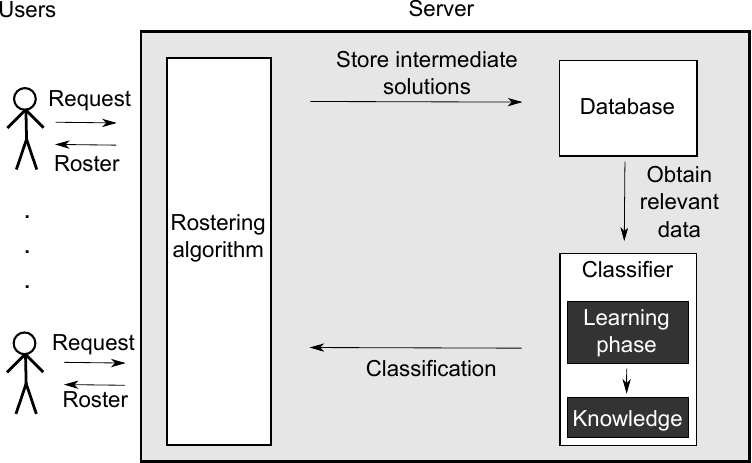}
\caption{\label{fig:motivation}Server-based rostering system}
\end{figure}

The motivation for this study is illustrated by the server-based rostering system shown in Figure~\ref{fig:motivation}. This server can solve many NRP or other related combinatorial problems at the same time. The users of the system submit their requests, i.e., tasks involving combinatorial problems, to the server where a rostering algorithm is running. A fast rostering algorithm is needed in order to minimize the response time of the server. In general, based on our experiments and statistics obtained by the Roster Booster tool \citep{srsolutions14}, rostering algorithms evaluate millions of rosters but most of them are very similar or obviously useless (bad quality). Moreover, the results obtained from the evaluation phase of the rostering algorithms are dropped practically immediately. Therefore, our approach includes two other components: a database and a classifier (learned knowledge). Using a database for storing the evaluated solutions allows the stored data to be employed to extract knowledge that facilitates the recognition of the quality of rosters much more rapidly. This knowledge can help to speed up the rostering algorithm when combined with the overall server application.

\subsection{Related Work}
The first studies that applied learning techniques to scheduling problems appeared in the 1990s.
A search-control policy for the resource-constrained scheduling problems was described by \cite{Zhang91}. A neural network is used as an approximator for a future value of the resource dilation factor in the next iteration of a scheduling algorithm. For each possible neighbor (i.e., all available changes from the current solution), the result obtained by the neural network is combined with an immediate reward value and the solution with the best evaluation is chosen. This approach, which was verified using the NASA space shuttle payload processing problem, outperformed the best known non-learning search algorithm at that time.

A neural network and logistic regression were used by \cite{Li11} to determine the correct order for executing low-level heuristics in a hyper-heuristic method. These techniques work as classifiers and they decide whether the execution order for low-level heuristics is good or not. Experiments based on exam timetabling problems showed that both methods could significantly speed up the original algorithm.

\cite{Li12} also recommended the use of pattern recognition to evaluate the quality of the solutions, where a neural network distinguishes good and bad solutions based on the structure of the overall solution. The theoretical results indicated that speedup was achieved for the NRP and the educational timetabling problem.

In general, neural networks are a suitable tool for pattern recognition \citep{Bishop95,Ripley08} because they can learn from experience and deal with noise in the input data almost as well as human beings. Case-based reasoning can also be employed for solving the NRP. \cite{Beddoe09} used this technique to repair constraint violations in the correct sequence, which can have a considerable impact on the roster quality.

In addition to learning techniques, other methods that focus on accelerating the evaluation process have been proposed previously. \emph{Delta evaluation} \citep{Ross94} is often applied so only the constraints affected by changes in a roster are evaluated instead of the whole set of constraints. This can significantly reduce the amount of time required, especially if the number of affected constraints is noticeably smaller than the whole set. \cite{Burke01} described a concept based on numbering, which maps a set of time units (i.e., the number of days times the number of shift types per day) onto a set of numbers. The ordered set is then utilized to evaluate the roster of an employee with the eligible constraints. 

The ideas outlined in the previous paragraph are definitely complementary to our approach. In particular, even if advanced techniques based on the delta evaluation are used, the evaluation process will still be more time consuming compared with classification (the detailed results and discussion are given in Section~\ref{sec:exp_ea}). Moreover, we consider difficult NRPs where complex constraints (see Section~\ref{sec:ps}) must be evaluated. The classifiers can handle these problems but advanced techniques to allow faster evaluations can rarely be used because there is no capacity for any reduction in practice, and these constraints require additional modifications in the source code. Therefore, a combination of those two approaches, i.e., classifiers and advanced techniques, might be very powerful.

\subsection{Contributions and Outline}
Our study is inspired by the work of \cite{Li12} where the authors made an advance in the alternative evaluation of the roster quality using machine learning in order to speed up its computation. The paper presents novel ideas, but it does not seem to provide a description of how machine learning should be integrated into a scheduling algorithm. The proposed technique also assumes fixed parameters of a problem which may cause a complete change in a classifier structure if the parameters are changed (e.g. a new nurse is hired). 

We provide a new view on this approach in order to make the technique usable in practice. The main contribution of our study is the design of a classifier that significantly reduces the time required by classical cost-oriented objective function evaluations. The classifier estimates whether or not a single change in the roster of one employee (nurse) improves the given objective function which results in a faster and more accurate neural-network learning. We suggest to use relative thresholds instead of one absolute threshold to determine whether a roster is good or bad. This is an important issue because the start of the rostering algorithm search process requires completely different thresholds compared with the end of the search. Since real-life problems are connected with the varying number of employees across different scheduling periods, our classifier is designed to handle this situation without additional learning from the new dataset or creating a new classifier. Furthermore, we demonstrate a possible application of the trained classifier to unknown data, based on the same or similar problem instances. Thus, the main improvements compared with \cite{Li12} can be summarized as follows.

\begin{itemize}
	\item We propose a more suitable application of a classifier to determine a roster's quality, i.e., determining whether a \textbf{change in the roster improves the value of the objective function} or not (see Section~\ref{sec:nnu}).
	\item Our method achieves \textbf{a better classification rate and speedup} (see Section~\ref{sec:exp}).
	\item \textbf{Our classifier is more robust}, e.g., it is not sensitive to changes in the number of employees to a certain extent (see Section~\ref{sec:nnu} and Section~\ref{sec:exp_p}).
	\item We illustrate the practical \textbf{use of the classifier in a tabu search algorithm} (see Sections~\ref{sec:ts} and \ref{sec:cts}).
\end{itemize}
The additional contributions of our study are as follows.
\begin{itemize}
	\item \textbf{The original design of weak classifiers} that can be used in boosting algorithms to achieve better quality rosters, but with a smaller speedup (see Section~\ref{sec:wc}).
	\item Our \textbf{classifier approach can speed up any (meta-)heuristic} (see Sections~\ref{sec:ts}, \ref{sec:nnu} and \ref{sec:cts}).
	\item The use of our solution in a \textbf{server-based rostering system helps to achieve a better response time} (see Section~\ref{sec:m} and Figure~\ref{fig:motivation}).
\end{itemize}

Finally, the speedup obtained can also be used to further exploit a larger unknown solution space in the NRP, thereby potentially finding a better roster. For example, with a speedup of four times, a rostering algorithm can perform at least three additional diversification processes but its runtime will be equal (or similar) to the same rostering algorithm without a classifier. However, this is not a major aspect of our study because we focus on achieving speedup, which is crucial for the response time of the server-based rostering algorithm depicted in Figure~\ref{fig:motivation}.

The remainder of this paper is organized as follows. The NRP is described in Subsection~\ref{sec:ps}. Subsection~\ref{sec:ts} presents a tabu search method as a simple rostering algorithm. Section \ref{sec:prfre} outlines the use of pattern recognition for roster evaluation, as well as describing a practical realization based on neural networks and boosting algorithms. In Section \ref{sec:exp}, we discuss the results obtained using our approach based on benchmark instances. Finally, we give our conclusions.

\section{Problem Statement}
\subsection{Nurse Rostering Problem}
\label{sec:ps}
In the case of the NRP, which is a specific version of the personnel rostering problem, activities are shifts with a defined start time, duration, and finish time. The resources are nurses who have no more than one shift assigned each day, and only if they have the necessary qualifications for that shift \citep{Dowsland98,Burke04}.

The problem is parameterized by the number of nurses $n$, the number of days in the planning period $d$, and the number of shifts $s$. Then, the roster $R$ is a three-dimensional binary matrix such that $\forall i \in \{1, \dots ,n\}, \forall j \in \{1, \dots ,d\}, \forall k \in \{1, \dots ,s\}$

\[
R_{ijk} = \begin{dcases*}
        1  & \parbox{6cm}{shift $k$ is assigned to nurse $i$ on day $j$,}\\
        0 & otherwise.
        \end{dcases*}
\]
The quality of roster $R$ is given by the objective function $Z$, which is defined as $$Z(R) = \sum_{i=1}^n Z_i(R),$$ where $Z_i$ is the quality of the assignment related to nurse $i$. The number of shifts that need to be assigned on a given day is called the coverage constraint.

\begin{figure}[ht]
\centering
\includegraphics[width=\linewidth]{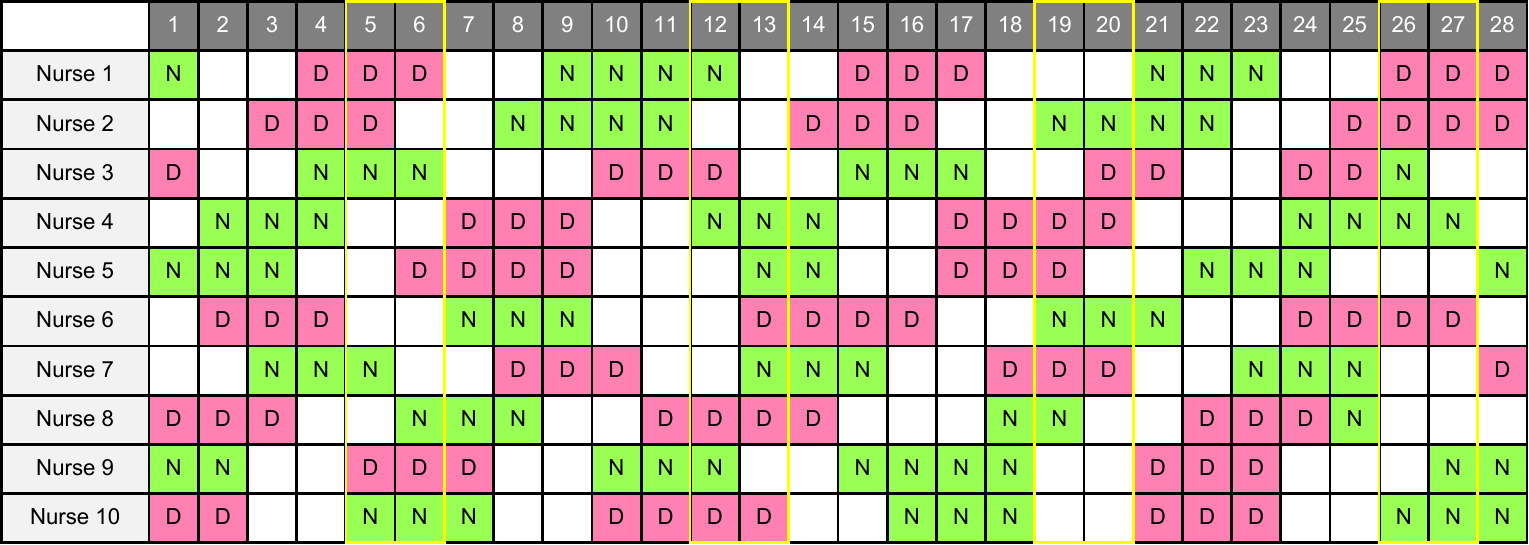}
\caption{\label{fig:roster}An example of a roster: 10 nurses, planning period of 28 days, and two shift types (day shift (D): 07:00--15:00; and night shift (N): 15:00--23:00).}
\end{figure}

Figure~\ref{fig:roster} shows an example of a simple roster for 28 days. In this case, the task was to assign two shift types (i.e., D = day shift and N = night shift) to 10 nurses. The desired coverage of shifts was three D shifts and three N shifts each day during the entire planning period.

During the gradual allocation of shifts to the roster, various constraints must be considered, which may refer to any roster dimension (days and nurses) or the shifts themselves. In general, there are two types of constraints: hard and soft. Hard constraints should never be violated and they have no effect on the value of the objective function $Z$. If the roster satisfies all of the hard constraints, it is a feasible solution to the problem; otherwise, it is an infeasible roster. In contrast to hard constraints, soft constraints may be violated. However, our goal is to achieve the minimum number of violations while considering the cost (weight) of a violation, which is reflected in the objective function $Z$. Thus, rostering requires the solution of an optimization problem where the minimum value of the objective function $Z$ must be found.

The type and number of hard and soft constraints depend on the problem instance. In this study, we consider the general NRP instances, which can be defined in the XML data format \citep{srsolutions15}. Thus, we do not focus on one specific group of instances (i.e., the same set of constraints and the problem has a similar structure). Nevertheless, to understand this problem better, the following list provides examples of \textbf{some} constraints that are often used, or which are interesting in terms of complexity.
\clearpage
\begin{itemize}
	\item Hard constraints:
	\begin{itemize}
		\item a nurse cannot be assigned to more than one shift per day,
		\item a shift requiring a particular skill cannot be covered by a nurse without that skill,
		\item the minimum time gap between consecutive shifts must be maintained (e.g., 12 hours),
		\item fixed shift/day off assignments must be preserved (e.g., a planned holiday of an employee),
		\item the required coverage of shifts for a day must be fulfilled (e.g., if there is a need for five early shifts on Monday, then exactly five nurses must be assigned to these five shifts).
	\end{itemize}
	\item Soft constraints:
	\begin{itemize}
	\item pattern-based (i.e., checking the minimum/maximum number of matches for the particular sequences of shift/day off (patterns) between any two dates in the planning period):		
		\begin{itemize}
			\item a nurse should have at least one free Sunday per planning period,
			\item a nurse should not work five or more consecutive days,
			\item a nurse should have at least 10 days off.
		\end{itemize}
	\item block-based \citep{baumelt10}, i.e., a block is a sequence of consecutive shifts in a schedule that meets some requirements:		
		\begin{itemize}
			\item the maximum number of working hours in a block should be equal to or less than the predefined number,
			\item the time gap between two consecutive blocks should be at least the predefined number: the so called ``minimal block rest'',
			\item the time gap between two consecutive shifts in a block should not exceed the ``minimal block rest''.
		\end{itemize}
	\item others:		
		\begin{itemize}
			\item working hours should be balanced with respect to the nurse's workload (e.g., overtime hours should be reduced to a minimum),
			\item nurses should have a similar number of shift types (e.g., to avoid a situation where one nurse only has night shifts and another only has early shifts, if this is not requested).
		\end{itemize}
	\end{itemize}
\end{itemize}	

Each constraint requires a different amount of time when it is evaluated in the cost-oriented objective function. From the viewpoint of complexity, the block-based constraints are the most difficult and time demanding, but an appreciable amount of time is also consumed by pattern-based constraints. In general, some constraints are very easy to evaluate, e.g., balancing a nurse's workload. In this case, the evaluation process is quite straightforward because it only accumulates the total workload of the given nurse's roster, and thus the time complexity is $O\left(n\right)$. However, for block-based and the pattern-based constraints, the evaluation procedure requires the updating of several auxiliary data structures for each single day in the roster of a nurse, as well as other processes. Thus, the time complexity is $O\left(m\cdot n\right)$, where $m$ is the length of the pattern. Furthermore, the evaluation procedure is often more complicated, where it contains several nested if-else constructions, which prevents the compiler from using the advanced capabilities of the CPU, such as vectorization. Therefore, we can see that the differences may be significant, and thus it would be very beneficial to develop an approach that eliminates/reduces those differences, thereby decreasing the overall time needed for the evaluation. This approach is presented in the following, specifically in Section \ref{sec:prfre}.

In the present study, we focus on instances with \textbf{complex constraints}, i.e., block-based and pattern-based constraints. Thus, real-world instances are employed to demonstrate the ability to handle complex constraints effectively and the standard benchmark instances are used to verify/compare the results obtained. Further details are provided in Section \ref{sec:exp}.
	
\subsection{Tabu Search}
\label{sec:ts}
One of the most successful meta-heuristics for NRP is tabu search (TS), which belongs to the class of local search algorithms \citep{glover98}. These algorithms search for a new solution among the neighbors of the current solution. They retain the best solution and continue until some stopping criterion is met (e.g., a time limit or no improvement after several steps). In general, the main problem that affects local search algorithms is their tendency to become stuck in a local extremum. TS partially resolves this issue by using an adaptive memory called the tabu list (of size $m$), which stores the characteristics of recently visited solutions to avoid using moves that lead to already visited solutions in the next $m$ iterations of the algorithm.

\begin{figure}[ht]
\begin{algorithmic}[1]
\REQUIRE initial solution $R$
\STATE $R_{best} = R_{actual} = R$
\STATE $best\_Z = actual\_Z = Z(R)$
\WHILE{stopping criterion is not met}
		\STATE select employee $emp_a$ for improvement such that the contribution of $emp_a$ to the whole $Z$ is the biggest and $emp_a$ is not forbidden by the terminating mechanism (i.e., $termination\_info[emp_a].iteration > size(tabu\_list)$)
    \IF{$emp_a = null$}
    	\STATE \textbf{break}	
    \ENDIF
    \STATE $local\_best\_Z = \infty$		
    \FOR{$\forall cand \in neighborhood(emp_a,R_{actual})$}
	    \IF{$cand$ is tabu}
	    	\STATE \textbf{continue}	
	    \ENDIF	
	    \STATE $R_{cand} =$ apply changes from $cand$ to the roster $R_{actual}$
	    \STATE /*calculate the value of $Z$ by only considering the modified employees */
    	\STATE $actual\_Z = actual\_Z + (Z_{cand.emp_a}(R_{cand}) - Z_{cand.emp_a}(R_{actual})) + (Z_{cand.emp_b}(R_{cand}) - Z_{cand.emp_b}(R_{actual}))$
	    \IF{$actual\_Z < local\_best\_Z$}
	    	\STATE $R_{local\_best} = R_{cand}$
	    	\STATE $local\_best\_Z = actual\_Z$
	    	\STATE $best\_candidate = cand$
	    \ENDIF     
    \ENDFOR
    \STATE $R_{actual} = R_{local\_best}$
    \STATE $actual\_Z = local\_best\_Z$
		\IF{$local\_best\_Z < best\_Z$}
			\STATE $best\_Z = local\_best\_Z$
		  \STATE $R_{best} = R_{local\_best}$
		  \STATE reset $termination\_info$
		  \STATE $tabu\_list.add(best\_candidate)$ 
		\ELSE
			\IF{$local\_best\_Z < termination\_info[emp_a].roster\_penalty$}
				\STATE $termination\_info[emp_a].roster\_penalty = local\_best\_Z$
				\STATE $termination\_info[emp_a].iteration = 0$
			\ELSE
				\STATE $termination\_info[emp_a].iteration++$
			\ENDIF
			\STATE $tabu\_list.add(empty\_record)$
		\ENDIF
\ENDWHILE
\RETURN $R_{best}$
\end{algorithmic}
\caption{\label{fig:ts1}Pseudo-code of the simple tabu search algorithm with a cost-oriented objective function evaluation}
\end{figure}

Figure~\ref{fig:ts1} shows the pseudo-code of the simple TS algorithm (TSA) considered in this study, which proceeds through the main loop with a stopping criterion, i.e., in our case, this is the number of iterations without any improvement in the objective function value $Z(R_{best})$. During each iteration of the main loop (lines 3--36), some employee $emp_a$ is chosen based on their contribution to $Z$, i.e., the algorithm tries to repair the roster for the employees with the worst shift assignments. Next, the algorithm enumerates all of the possible exchanges of the shifts in the roster where one of the affected employees is employee $emp_a$, i.e., the current solution neighborhood with respect to employee $emp_a$ (see line 9). An individual exchange is expressed as a triple:
\begin{itemize}
	\item $cand = \left\langle emp_a, emp_b, day\right\rangle$, which represents an exchange of two different shifts (or one shift and one day off) in the roster between employees $cand.emp_a$ and $cand.emp_b$ on the specific \textit{day}. Then, the tabu list stores a tuple $tl_{cand} = \left\langle emp_a, emp_b, shift_a, shift_b, day\right\rangle$
	\item or $cand = \left\langle emp_a, day, shift\right\rangle$, which represents the replacement of a shift (or day off) by a new shift (or day off) in the roster of employee $cand.emp_a$ on the specific \textit{day}. In this case, the tabu list stores a tuple $tl_{cand} = \left\langle emp_a, shift_{old}, shift_{new}, day\right\rangle$.
\end{itemize}
The candidate evaluations (see line 15) are calculated only for the modified employees and not for the entire roster, which greatly increases the speed of the computation. Moreover, the delta evaluation is used, i.e., only the affected constraints/part(s) of constraints are considered. Nevertheless, the cost-oriented objective function evaluation is still a critical task in terms of time complexity. The algorithm typically spends 80\% of its whole runtime processing this part of the code, based on our measurements and experiences with other algorithms. Furthermore, excessive numbers of obviously bad solutions are evaluated unnecessarily during the runtime of the TSA and other algorithms. Therefore, the use of a technique that can rapidly determine whether the structure of the roster is bad or good would significantly speed up the execution of the TSA and other approaches. 

\section{Pattern Recognition during Roster Evaluation}
\label{sec:prfre}
In the NRP, the evaluation of the partial solutions requires much of the computational time. The core idea of our approach is to speed up meta-heuristics to solve the NRP by using classifiers that are pattern recognition techniques \citep{chen10}. The classifier aims to assign each input vector (pattern) to the correct class according to the experience gained in the learning phase, based on the training data. The typical classifier realization is a neural network.

\subsection{Background: Neural Networks}
\label{sec:bnn}
Many types of neural networks exist, but we consider a multilayer perceptron network because it is one of the most widely used approaches and it obtained the best general results in our experiments. The basic components of a neural network are neurons, which are arranged into layers and connected via links. Each link $(u,v)$ has its weight $w_{uv}$, where $u$ is the source neuron and $v$ is the target neuron. The output $y_v$ of the neuron $v$ is defined as
$$
	y_v = F\left(\sum_{\forall u|(u,v) \in \delta^{-}(v)} w_{uv} \cdot y_u+\Theta_v\right),
$$
where $\delta^{-}(v)$ is a set of links entering the neuron $v$, $\Theta_v$ is a threshold of the neuron $v$, and $F$ is an activation function (e.g., the sigmoid function).
To train a neural network, training data are needed in the following format: (input pattern, corresponding class). The training (learning) process continuously updates the weights of the links to improve the behavior of the neural network based on the input data. Thus, the result produced by the neural network should be the same as the desired output for each input pattern. It is also necessary to use test data to avoid the problem of overfitting; otherwise, the trained network would not be applicable to data other than the training set.

\begin{figure}[ht]
\centering
\includegraphics[width=8cm]{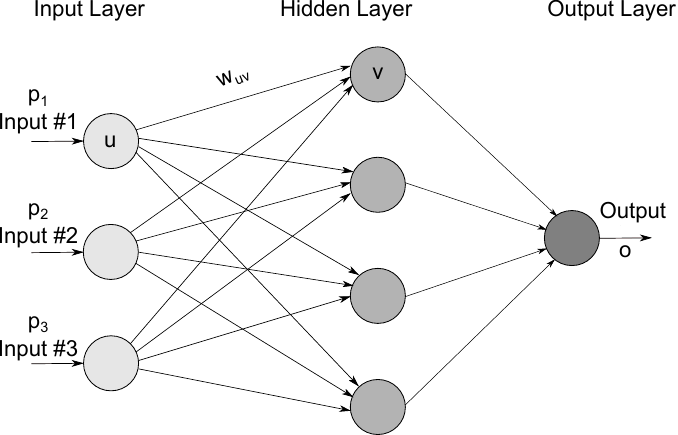}
\caption{Example of a neural network}
\label{fig:ann}
\end{figure}

A two-layer neural network, i.e., multilayer perceptron network, is shown in Figure~\ref{fig:ann}.\footnote{In this study, we do not count the input layer as a layer.} There are three input nodes, one output neuron, and four hidden neurons. The use of one hidden layer allows us to reproduce any differentiable function. By contrast, a network without hidden layers (i.e., a one-layer network) can only work with linearly separable problems \citep{yegnanarayana09}.

\subsection{Use of a Neural Network for Roster Evaluations}
\label{sec:nnu}

\begin{figure}[ht]
\centering
\includegraphics[width=10cm]{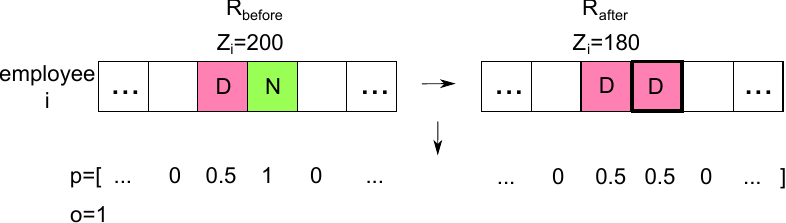}
\caption{A change in the roster for employee $i$}
\label{fig:pattern}
\end{figure}

In our approach, we do not use the absolute value of the objective function $Z$ to determine whether a roster is good or bad (as in \cite{Li12}). Instead, we are interested in the relative changes in $Z$ from roster $R_{before}$ to $R_{after}$. Therefore, the input pattern entered into the neural network is treated as a simple change in the roster of one employee (see Figure~\ref{fig:pattern}). The input pattern is regarded as a vector $p$ of length $2 \cdot d$, where $d$ is the number of days in the planning period. Its element $p_r, \forall r~\in~\{1,\dots,d\}$ represents the roster of employee $i$ before the change and $p_r, \forall r \in \{d+1,\dots,2 \cdot d\}$ is that after the change. Each element of vector $p$ has a value from the interval $[0,1]$, where $0$ denotes a day off and the other values represent shifts. The encoded value of a shift $k$ is equal to $k \cdot \frac{1}{s}$, where $s$ is the number of shifts and $k~\in~\{1, \dots ,s\}$. The output $o$ of the neural network expresses whether the change leads to an improvement in the objective function or not. The input pattern represents a change in the roster, so vector $p$ is classified as good ($o=1$) if $(Z_i(R_{before}) - Z_i(R_{after})) > 0$, and as bad ($o=0$) if $(Z_i(R_{before}) - Z_i(R_{after})) \leq 0$.

\cite{Li12} use one entire roster as a single large pattern, whereas our approach considers the relative changes in the roster described in the previous paragraph, thereby making evaluations more flexible and faster. Nevertheless, the results obtained are not satisfactory even with this major improvement because the algorithm needs to classify two exchanges (the rosters of two employees are changed; see Figure \ref{sec:ts}, line 14). The main problem occurs when one of the exchanges is marked as good and the second as bad, where the algorithm is not able to resolve this ``draw''. The second problem occurs with equally classified exchanges because it is impossible to decide which of the pair is better. 

These drawbacks can be eliminated by replacing two classes (i.e., good and bad) with five ``pseudo-classes'' (very good, good, equal, bad, and very bad). The ranges for improving the objective function $(Z_i(R_{before}) - Z_i(R_{after}))$ are determined with respect to the considered constraints and their weights:
\begin{itemize}
	\item very good (desired output $o=1$): $(Z_i(R_{before}) - Z_i(R_{after})) >> 0$,
	\item good ($o=0.7$): $(Z_i(R_{before}) - Z_i(R_{after})) > 0$,
	\item equal ($o=0.5$): $(Z_i(R_{before}) - Z_i(R_{after})) = 0$,
	\item bad ($o=0.3$): $(Z_i(R_{before}) - Z_i(R_{after})) < 0$,
	\item very bad ($o=0$): $(Z_i(R_{before}) - Z_i(R_{after})) << 0$.
\end{itemize}
The aim of this step is to derive more precise information from the classifier, which can then be used in a more accurate decision process. In addition, this approach improves the classification rate (i.e., the percentage of successfully classified data) because the approximation of objective function $Z$ is better.

\subsection{Background: Boosting Algorithms}
\label{sec:ba}
Neural networks, especially multi-layer networks, can achieve a good classification rate, but the rate is not sufficient in some situations. In these cases, boosting algorithms \citep{buhlmann07} can be used to achieve a better classification rate. In general, these algorithms combine weak classifiers to create a strong classifier (where the output is the weighted sum of the outputs of the weak classifiers). The weights of the weak classifiers are adjusted during the learning phase of the boosting algorithm. The weak classifiers should be simple in terms of the classification speed, e.g., one-layer neural networks or small decision trees; otherwise, the runtime reduction for the rostering algorithm would be negligible with such a weak classifier.

A well-known boosting algorithm is Adaptive boosting (AdaBoost), which is a meta-algorithm that can improve the performance of other learning algorithms \citep{freund95}. The simple weak classifiers $h_t$ are combined together to create a more powerful strong classifier $H$ called a committee.

The classification of input pattern $p$ is given as:
$$H(p) = sign\left(\sum_{t=1}^T \alpha_t h_t(p)\right),$$

\noindent
where $T$ is the number of weak classifiers and $\alpha$ is explained later in the text. After each weak classifier is found, the input patterns
$p_i$ (with labels/outputs $o_i \in \left\{-1,1\right\}$) are re-weighted in their distribution $D_t(i)$, based on the current classifier performance. In the case of AdaBoost, the misclassification error is punished with a negative
exponential cost function such that $D_0(i) = \frac{1}{\left|P\right|}$, and the weight for step t + 1 is given as:
$$D_{t+1}(i) = \frac{D_t(i)exp(-\alpha_t o_i h_t(p_i))}{L_t},$$

\noindent
where $L_t$ is a normalization factor, which is selected such that $D_{t+1}$ is a probability distribution. Therefore, this error is minimized, which gives us an analytical solution to parameter $\alpha_t$:
$$\alpha_t=\frac{1}{2}ln\left(\frac{1+r}{1-r}\right)\mbox{, where } r=\sum_i D_t(i) o_i h_t(p_i).$$

The main disadvantage of the AdaBoost algorithm is its classification speed because all of the weak classifiers must classify the input pattern. As mentioned earlier, the first weak classifiers can decide some easier input patterns, which can be accomplished by a ``classifier cascade,'' i.e., the concatenation of several weak classifiers where the next weak classifier in the cascade uses the output from the previous weak classifier (some kind of voting system). The first use of a ``classifier cascade'' was proposed in a sequential probability ratio test framework by \cite{wald45} for constructing the WaldBoost detector in \cite{sochman05}.

\subsection{Weak Classifiers for Roster Evaluation}
\label{sec:wc}
In this study, we also consider boosting algorithms, i.e., AdaBoost and WaldBoost, which include dozens of weak classifiers for the roster evaluation. These weak classifiers differ according to two features. The first feature is the shift assignment representation in the input pattern $p$, i.e.:
\begin{itemize}
	\item the shift is represented as a real number -- e.g., a day off is represented as $0$, a day shift as $0.5$, and a night shift as $1$ (used in Section \ref{sec:nnu});
	\item binary encoding -- defined by $R_{ijk}$, i.e., one input of the classifier is a bit ($j \cdot s + k$), which indicates whether shift $k$ is assigned to employee $i$ on day $j$ (e.g., a day off is represented as ``0 0 0'', a day shift as ``0 1 0'', and a night shift as ``0 0 1'').
\end{itemize}

Previous encodings are applied to the entire planning period, but this may be unnecessary in some cases because only a small part of the planning period is changed at once. Therefore, based on the aforementioned representation, it may be beneficial to considering only a small neighborhood with $x$ days. 

The second feature uses additional information derived from the structure of the roster for an employee. This additional information is appended to the input pattern $p$, i.e., the length of vector $p$ is increased by one for each piece of additional information because the information is expressed as a single number. The abbreviations used in the example have the following meanings: O = day off, D = day shift, N = night shift, and * = any type of shift. The additional information is:
\begin{itemize}
	\item the number of isolated shifts (e.g., ``... O * O ...''), days off (e.g., ``... * O * ...''), or both;
	\item the maximal length of a block containing shifts or days off, e.g., for the roster ``O O O D D D O O N N N N,'' it is 4 in the case of the shifts and 3 in the case of the days off; 
	\item the number of shifts assigned;
	\item the number of blocks of shifts or days off, e.g., for the roster ``O O D O O N O O N N,'' it is 3 in both cases, 
	\item the number of transitions between shift types, e.g., roster ``O D D D D'' has one transition whereas ``O D D N D'' has three transitions.
\end{itemize}
Binary encoding is the most suitable for neural networks because the weights are either propagated further or zeroed (see Section \ref{sec:bnn}). Thus, this network can work more precisely than that with real numbers, where the weights are distributed with some coefficient based on the shift. However, the input pattern $p$ can be quite large with a large number of shifts and a long planning period. Therefore, neural network training is much slower as well as the classification process.
Additional information cannot be used separately because it has very low interpretive value in terms of the entire situation for the roster of an employee. Therefore, this information is used as an addition to the shift assignment representation. The influence is noticeable but the impact of this change is not crucial in general. 

In the following, we do not distinguish between a simple classifier and a complex classifier (i.e., from the boosting algorithm). If necessary, we use the phrases ``simple classifier'' or ``complex classifier'' explicitly.

\subsection{Classifier in Tabu Search Algorithm}
\label{sec:cts}
In general, the trained classifier will never have a 100\% success rate when matching the input pattern to the correct class. Moreover, the algorithm must classify two patterns, and thus the total success rate is even slightly worse (see Section \ref{sec:nnu}). Therefore, we use the classifier as a filter in the TSA, which eliminates the majority of the potentially bad solutions. The remaining solutions are then evaluated by the cost-oriented objective function $Z$. If the TSA depends only on the classifier, it would easily become stuck in a place where the classifier fails systematically. The code in Figure~\ref{fig:ts2} represents the inner loop of the TSA from Figure~\ref{fig:ts1} (lines 8--21), which has been extended with the filter based on a classifier. The filter stores a specific number of candidate solutions (given by $filter\_size$) with the highest $classification$ (lines 7--12 in Figure~\ref{fig:ts2}). The list is subsequently processed by the cost-oriented evaluation method (lines 14--22 in Figure~\ref{fig:ts2}).

\begin{figure}[ht]
\begin{algorithmic}[1]
		\STATE $filter\_size = 50$ /*number related to filter elimination*/
		\STATE $list\_cand = null$ /*list of best candidates from the viewpoint of the classifier */
    \FOR{$\forall cand \in neighborhood(emp_a,R_{actual})$}
	    \IF{$cand$ is tabu}
	    	\STATE \textbf{continue}	
	    \ENDIF	
    	\STATE $classification = classifier(cand.emp_a) + classifier(cand.emp_b)$
    	\IF{$size(list\_cand) < filter\_size$}
    		\STATE $list\_cand.add(cand)$  
	    \ELSIF{$worst\_classification(list\_cand) < classification$}
	    	\STATE $list\_cand.replace\_worst(cand)$ 
	    \ENDIF   
    \ENDFOR
    \FOR{$\forall cand \in list\_cand$}
    	\STATE $R_{tmp} =$ apply changes from $cand$
    	\STATE $tmp\_Z = Z(R_{tmp})$
	    \IF{$tmp\_Z < local\_best\_Z$}
	    	\STATE $R_{local\_best} = R_{tmp}$
	    	\STATE $local\_best\_Z = tmp\_Z$
	    	\STATE $best\_candidate = cand$
	    \ENDIF   
    \ENDFOR
\end{algorithmic}
\caption{\label{fig:ts2}Pseudo-code of the modified tabu search algorithm with a classifier as a filter. This code replaces that in lines 9--21 from Figure \ref{fig:ts1}.}
\end{figure}

\section{Experiments}
\label{sec:exp}
In this section, we present the results of experiments obtained using standard \citep{curtois13} and real-world benchmark instances \citep{baumelt10}. First, we describe the experimental setup. In Subsection \ref{sec:exp_o}, we compare the different learning methods used for classifier-based evaluation. Next, we explain the differences among the various evaluation approaches. In Subsection \ref{sec:exp_p}, we then compare three approaches to show how they differ in terms of the evaluations of the intermediate solutions during the runtime of the rostering algorithm. These approaches are:
\begin{enumerate}
	\item standard cost-oriented evaluation, i.e., evaluation of soft constraints (see Section \ref{sec:ts});
	\item simple classifier, i.e., the neural network described in Section \ref{sec:nnu};
	\item complex classifiers, i.e., boosting algorithms using the weak classifiers described in Sections \ref{sec:ba} and \ref{sec:wc}.
\end{enumerate}
These experiments are focused on the speed of the three approaches and the quality of the resulting rosters. Note that speed is a very important factor in this study because we use the algorithm in a server-based rostering system, where many users submit their requests at the same time, and thus the system response time should be kept as low as possible. Finally, Subsections~\ref{sec:disc} and \ref{sec:comp} conclude this section by comparing and discussing the results obtained.

\subsection{Experimental Setup}

The proposed approach was verified on a computer with an Intel Core i7-3520M 2.90 GHz CPU and 8.0 GB DDR3 RAM. Our TSA was implemented in C\#. The experiments were executed using the standard benchmark instances: ``Valouxis,'' ``Millar,'' ``Millar-s,'' ``Ortec,'' and ``GPost'' \citep{curtois13}, and the real-world benchmark instances: ``bp01,'' ``bp02,'' ``bp03,'' ``bp04,'' and ``bp05'' \citep{baumelt10}. These specific standard instances were chosen to cover a wide spectrum of instance types. The real-world instances were inspired by a real-world problem and they were selected to demonstrate the efficiency of our approach with a much larger input data set and more complex soft constraints. Descriptions of the instances are summarized in Table~\ref{tab:desc}, where the last column ``RBC'' indicates the complexity of the instance computed by the tool Roster Booster \citep{srsolutions14}.

\begin{table}[!htb]
\centering
\caption{\label{tab:desc}Parameters of the test instances}
\begin{tabular}{|l|r|r|r|r|r|}
\hline
\multirow{2}{*}{Instance} & \multicolumn{4}{|c|}{Number of} & \multirow{2}{*}{RBC}\\
\cline{2-5}
& shift types & days & employees & constraints & \\
\hline\hline
Millar & 2 & 14 & 8 & 14 & $10^{53}$\\
Millar-s & 2 & 14 & 8 & 8 & $10^{53}$\\
Ortec & 4 & 31 & 16 & 52 & $10^{347}$\\
Valouxis & 3 & 28 & 16 & 18 & $10^{270}$\\
GPost & 2 & 28 & 8 & 33 & $10^{107}$\\
\hline\hline
bp01 & 102 & 28 & 86 & 14 & $10^{7270}$\\
bp02 & 96 & 28 & 88 & 61 & $10^{7343}$\\
bp03 & 124 & 35 & 88 & 77 & $10^{9042}$\\
bp04 & 118 & 28 & 90 & 38 & $10^{7846}$\\
bp05 & 79 & 35 & 86 & 15 & $10^{8020}$\\
\hline
\end{tabular} 
\end{table} 

In our experiments, we only used two-layer neural networks as the simple classifiers because the neural network without a hidden layer did not achieve a good rate. In addition, using more than two layers did not achieve further improvements, i.e., the slightly better classification rate did not justify the slower classification process and the learning phase for the neural network. Each neuron used a sigmoid activation function and the hidden layer comprised 10 neurons.

For each benchmark instance, the classifier was trained using data that was obtained as follows: we collected all of the intermediate solutions from 500 TSA runs, where each run had a different initial solution and generated unique data (which was verified in every experiment). Subsequently, 10000 different input patterns (samples) were equally distributed to the classes. Next, based on the best practices in the neural network domain, 70\% were used as the training data and 30\% as the test data. 

A back-propagation method \citep{rumelhart86} was used for neural network training with a learning rate of 0.3 and 100 iterations (epochs), which corresponded to approximately 2 seconds in these conditions. The test data set was used to detect overfitting of the trained classifier and to verify the classification rate. Boosting algorithms, AdaBoost and WaldBoost, were learned in 10 iterations, i.e., by considering the combination of 10 weak classifiers. The learning phase for the simple classifiers, AdaBoost, and WaldBoost, required approximately 0.1, 4, and 5 seconds, respectively, for one-layer neural networks, whereas two-layer neural networks required 1.5, 39, and 41 seconds, respectively, in the same conditions. 

The experiments with the TSA (see Figure~\ref{fig:ts2}) were performed using the initial solutions, which differed from those used in the learning phase. Moreover, in our experiments, \textbf{the TSA never generated a pattern that was used in the learning phase}.

\subsection{Comparison of the Learning Methods}
\label{sec:exp_o}

Table~\ref{tab:exp_lm} shows the effect of applying different learning methods in the TSA for five standard benchmark instances. We compared the neural network approach with logistic regression and decision trees. The column ``Class. rate'' stands for the classification rate, i.e., the success of the learning method based on its decisions regarding the test data. The fourth column denotes the total number of evaluations performed by the TSA. The next column represents the average time consumed per evaluation, using the given learning method. The last column shows the deviation in the quality of the solution obtained (i.e., the difference in the number of soft constraint violations, where + denotes a worse result). In the quality comparison, the roster obtained from the cost-oriented evaluation was used as the reference solution.

\begin{table}[!htb]
\catcode`\-=12
\centering
\caption{\label{tab:exp_lm}Experiments with the selected learning methods (NN: neural networks; LR: logistic regression; DT: decision trees)}
\begin{tabular}{|l|l|r|r|r|r|}
\hline
Instance & \parbox{1.5cm}{Learning method} & Class. rate [\%] & \# evaluations & \parbox{1.2cm}{time per eval. [$\mu$s]} & \parbox{1.7cm}{Difference in sol. quality} \\
\hline\hline
\multirow{3}{*}{Valouxis} & NN & 84.76 & 45293 & 8.71 & -1 \\
 & LR & 70.98 & 12582 & 21.73 & +10 \\
 & DT & 75.83 & 21475 & 982.49 & +8 \\
\hline
\multirow{3}{*}{Millar} & NN & 89.43 & 4112 & 5.19 & 0 \\
 & LR & 72.73 & 1593 & 12.53 & +6 \\
 & DT & 78.87 & 4981 & 492.57 & +4 \\
\hline
\multirow{3}{*}{Millar-s} & NN & 91.09 & 3492 & 5.14 & 0 \\
 & LR & 73.67 & 1269 & 12.61 & +3 \\
 & DT & 79.18 & 3193 & 478.13 & +2 \\
\hline
\multirow{3}{*}{Ortec} & NN & 85.22 & 158362 & 9.23 & +2 \\
 & LR & 70.13 & 93452 & 23.47 & +15 \\
 & DT & 74.39 & 118294 & 1086.74 & +13 \\
\hline
\multirow{3}{*}{Gpost} & NN & 85.15 & 26700 & 8.72 & +3 \\
 & LR & 71.17 & 22847 & 22.01 & +16 \\
 & DT & 76.28 & 45286 & 993.86 & +13 \\
\hline
\end{tabular} 
\end{table} 

The results in Table~\ref{tab:exp_lm} are quite unequivocal, where neural networks outperformed the other methods in terms of all the parameters. The poor quality of the solutions obtained by logistic regression and decision trees was caused by their inferior classification rates, which were well below 80\%. Moreover, these methods were significantly slower than the neural network-based approach, especially decision trees. Thus, based on the overall performance, \textbf{only the neural network-based method was used} in the more detailed experiments.

\subsection{Comparison of the Evaluation Approaches}
\label{sec:exp_ea}

The experiments presented in this subsection had two objectives: (i) to demonstrate the usability of our approach inside other (meta-)heuristics, and (ii) to compare classifier-based evaluation with delta-based evaluation methods. In addition to the TSA, other heuristics were considered, i.e., hill climbing and simulated annealing. Our proposed classifier evaluation was compared with delta evaluation, which has been used very frequently in previous studies, albeit with slightly different modified versions. The delta evaluation methods investigated were:
\begin{itemize}
	\item ``eval'' - a cost-oriented evaluation without any delta features, i.e., each time a change occurs, the entire roster is evaluated;
	\item ``$\Delta$ C'' - a cost-oriented evaluation where only the constraints that might be affected by a change are evaluated, although the entire roster is still evaluated;
	\item ``$\Delta$ E'' - a cost-oriented evaluation where only part of the roster is evaluated, which comprises all the nurses' rosters affected by a change; 
	\item ``$\Delta$ EC'' - a combination of ``$\Delta$ C'' and ``$\Delta$ E,'' i.e., only part of the roster is evaluated using the constraints affected by the change;
	\item ``$\Delta$ A'' - based on ``$\Delta$ EC'' but it goes even deeper, where the evaluation only considers the affected part of the nurses' roster that is decisive for the total penalty.
\end{itemize} 

Table~\ref{tab:exp_ep} shows the effects of the different evaluation methods on the selected heuristics. The third column represents the number of evaluations for each delta-based evaluation method. The subsequent columns stand for the average time consumed per evaluation for each delta-based evaluation method. The final columns present the previously described information for the classifier-based approach.

\begin{table}[!htb]
\catcode`\-=12
\centering
\caption{\label{tab:exp_ep}Experiments with the selected evaluation methods applied in different heuristics (HC: hill climbing; SA:  simulated annealing)}
\scalebox{0.81}{
\begin{tabular}{|l|l||r|r|r|r|r|r||r|r|r|}
\hline
\multirow{3}{*}{Instance} & \multirow{3}{*}{Heuristic} & \multicolumn{6}{|c||}{delta evaluation methods} & \multicolumn{3}{|c|}{\multirow{2}{*}{classifier}} \\
\cline{3-8}
& & \multirow{2}{*}{\# evals.} & $\Delta$ A & $\Delta$ EC & $\Delta$ E & $\Delta$ C & eval & \multicolumn{3}{|c|}{} \\
\cline{4-11}
& & & \parbox{0.6cm}{time [$\mu$s]} & \parbox{0.6cm}{time [$\mu$s]} & \parbox{0.6cm}{time [$\mu$s]} & \parbox{0.6cm}{time [$\mu$s]} & \parbox{0.6cm}{time [$\mu$s]} & \# evals. & \parbox{0.6cm}{time [$\mu$s]} & qual. \\
\hline\hline
\multirow{3}{*}{Valouxis} & TS & 32689 & 57.53 & 65.85 & 70.24 & 523.32 & 670.18 & 45293 & 8.71 & -1\\
 & HC & 27541 & 59.72 & 67.39 & 70.22 & 518.85 & 670.12 & 31256 & 8.70 & +1\\
 & SA & 28764 & 54.83 & 66.12 & 70.23 & 525.39 & 670.15 & 23567 & 8.71 & +2\\
 \hline
 \multirow{3}{*}{Millar} & TS & 2953 & 26.94 & 32.67 & 36.10 & 102.54 & 173.89 & 4112 & 5.19 & 0\\
 & HC & 1876 & 25.15 & 30.29 & 36.11 & 106.28 & 173.81 & 1934 & 5.18 & 0\\
 & SA & 2045 & 25.73 & 33.81 & 36.10 & 100.10 & 173.92 & 2123 & 5.18 & +1\\
 \hline
 \multirow{3}{*}{Millar-s} & TS & 4561 & 13.35 & 18.49 & 22.58 & 78.23 & 109.64 & 3492 & 5.14 & 0\\
 & HC & 1956 & 14.10 & 17.72 & 22.59 & 75.85 & 109.63 & 2314 & 5.13 & 0\\
 & SA & 2213 & 13.05 & 16.98 & 22.57 & 76.55 & 109.61 & 2654 & 5.11 & +1\\
 \hline
 \multirow{3}{*}{Ortec} & TS & 55046 & 98.69 & 101.83 & 113.02 & 923.51 & 1086.52 & 158362 & 9.23 & +2\\
 & HC & 47296 & 99.88 & 104.26 & 113.01 & 928.43 & 1086.51 & 63451 & 9.21 & +3\\
 & SA & 52849 & 94.14 & 103.11 & 113.01 & 925.74 & 1086.50 & 53958 & 9.22 & +5\\
 \hline
 \multirow{3}{*}{Gpost} & TS & 17347 & 61.88 & 69.75 & 75.96 & 674.82 & 711.84 & 26700 & 8.72 & +3\\
 & HC & 13517 & 59.34 & 66.97 & 75.95 & 671.59 & 711.79 & 14456 & 8.70 & +2\\
 & SA & 15382 & 58.09 & 68.25 & 75.94 & 672.16 & 711.86 & 17472 & 8.71 & +4\\
\hline
\end{tabular}
} 
\end{table} 

The evaluation methods related to delta evaluation are simple to compare. We can see that the total number of evaluations is the same in each row (i.e., one instance and one heuristic) because all of the methods are exact and they perform the same process, only in different ways. Thus, it is also a logical implication that the final roster quality is the same. Nevertheless, there were evident differences in the time needed for one evaluation. Obviously, from the definition, the ordering is ``eval'' $\geq$ ``$\Delta$ C'' $\geq$ ``$\Delta$ E'' $\geq$ ``$\Delta$ EC'' $\geq$ ``$\Delta$ A''. However, there was a very small difference between ``$\Delta$ E'' and ``$\Delta$ EC'' for some instances. This was because (almost) all of the constraints were always affected, regardless of the changes in the roster made by the algorithm. The difference between ``$\Delta$ EC'' and ``$\Delta$ A'' is noticeable, but it might be higher for some instances. This issue is connected with the complexity of the constraints, where the part of the nurse's roster that needs to be re-evaluated was excessively large, so it would have been more efficient to evaluate the entire nurse's roster. Moreover, the demand on the auxiliary code increased, which in the worst case yielded times very similar to ``$\Delta$ EC.''

The classifier evaluation achieved better results in terms of the overall speed, although it needed more evaluations for some instances, which was due to the significantly shorter time consumed per evaluation. In addition, the quality obtained was very similar or slightly worse for some instances. We should also highlight the fact that the classifier evaluation required nearly the same time for one evaluation in each heuristic and instance with the same number of days. This demonstrates that our proposed approach is indifferent to the number and complexity of constraints (provided that the neural network can successfully learn the training set for the given problem). In contrast to the classifier approach, the other evaluation methods required different times for the instances because they differed in terms of the number and complexity of the constraints, as well as in the number of days. The times required by the heuristics were more or less equal for the same instances. The small differences were caused by the method employed by each heuristic to search the state space, and thus each heuristic evaluated various intermediate solutions.  

In the remaining experiments, \textbf{``$\Delta$ EC'' was used as a cost-oriented evaluation} method because it was the second best evaluation method among all of the methods considered, and it had significant advantages from an implementation perspective. Furthermore, the difference between ``$\Delta$ EC'' and ``$\Delta$ A'' was negligible for the more difficult real-world benchmark instances due to the complexity of the constraints. Finally, \textbf{the TSA was used as the reference heuristic} in further experiments because it produced rosters with overall better quality (see Table~\ref{tab:exp_ep}).

\subsection{Performance of the Rostering Algorithm with Different Evaluation Approaches}
\label{sec:exp_p}

Table~\ref{tab:exp} shows the most important results obtained after comparing the cost-oriented objective function evaluation and using the simple classifier as a filter, which eliminated 90\% of the potentially bad solutions during each iteration of the TSA. The first column identifies the benchmark instance. The next two columns show the CPU time consumed by the classifier-oriented (where the achieved speedup is indicated in the brackets) and the standard cost-oriented evaluation, respectively. Only two parts of the entire roster, i.e., the roster for employee $cand.emp_a$ and $cand.emp_b$, were re-evaluated (``$\Delta$ EC'' evaluation). To measure the results as accurately as possible, we only counted the total time required by the given evaluation methods. Finally, the last two columns stand for the misclassification rates (i.e., rates of incorrectly rejected solutions and accepted solutions, respectively). 

\begin{table}[!htb]
\catcode`\-=12
\centering
\caption{\label{tab:exp}Results of experiments using the trained simple classifiers (neural network)}
\begin{tabular}{|l|r|r|r|r|r|}
\hline
\multirow{2}{*}{Instance}  & \multicolumn{2}{|c|}{CPU time [s]} & \multirow{2}{*}{\parbox{3cm}{Difference in solution quality}} & \multicolumn{2}{|c|}{Misclassification [\%]}\\
\cline{2-3}
\cline{5-6}
& classifier & standard & & reject & accept \\
\hline\hline
Valouxis & 0.39 (9.9x) & 3.86 & -1 & 9.5 & 15.7 \\
Millar & 0.03 (3.7x) & 0.11 & 0 & 6.2 & 13.6 \\
Millar-s & 0.02 (3.5x) & 0.07 & 0 & 5.6 & 13.5 \\
Ortec & 1.89 (2.8x) & 5.23 & +2 & 10.6 & 16.3 \\
Gpost & 0.38 (2.8x) & 1.07 & +3 & 7.8 & 15.1 \\
\hline\hline
Average & 0.54 (3.8x) & 2.07 & +0.8 & 7.9 & 14.8 \\
\hline\hline
bp01 & 98 (7.4x) & 721 & +7 & 9.1 & 14.4 \\
bp02 & 119 (7.0x) & 833 & +9 & 9.2 & 15.9 \\
bp03 & 182 (6.1x) & 1110 & +14 & 10.7 & 17.8 \\
bp04 & 129 (6.9x) & 890 & +11 & 9.2 & 16.7 \\
bp05 & 89 (7.8x) & 693 & +8 & 8.3 & 14.3 \\
\hline\hline
Average & 123.4 (6.9x) & 849.4 & +9.8 & 9.3 & 15.8 \\
\hline
\end{tabular} 
\end{table} 

Table~\ref{tab:exp_modif} shows the behavior of the simple classifier when the problem instance was changed slightly. The particular modification of the instance is indicated in the brackets next to its name: 
\begin{itemize}
	\item[a)] \textit{sc+/sc-} -- one soft constraint added (i.e., ``forbidden consecutive shifts'': day | evening | night)/removed (i.e., ``forbidden isolated shifts'');
	\item[b)] \textit{3sc+/3sc-} -- three soft constraints added (i.e., ``forbidden consecutive shifts'': day | evening | night, ``the maximum number (4) of consecutive day shifts'' and ``at least two free weekends'')/removed (i.e., ``forbidden isolated shifts,'' ``forbidden consecutive shifts'': night | evening and ``forbidden consecutive shifts'': evening | night);
	\item[c)] \textit{e+/e-} -- one employee added/removed;
	\item[d)] \textit{3e+/3e-} -- three employees added/removed.
\end{itemize}

For example, ``Valouxis (sc+)'' is the standard Valouxis instance with one soft constraint added. The same classifier was used for each corresponding instance, e.g., for Valouxis in Table~\ref{tab:exp} and Valouxis (sc-) in Table~\ref{tab:exp_modif}, we considered the same weights in the multilayer perceptron network.

\begin{table}[!htb]
\catcode`\-=12
\centering
\caption{\label{tab:exp_modif}Results of experiments with the trained simple classifiers (neural network) for similar instances}
\begin{tabular}{|l|r|r|r|}
\hline
\multirow{2}{*}{Instance}  & \multicolumn{2}{|c|}{CPU time [s]} & \multirow{2}{*}{\parbox{3cm}{Difference in solution quality}} \\
\cline{2-3}
& classifier & standard &  \\
\hline\hline
Valouxis (sc-) & 0.21 (11.2x) & 2.36 & 0 \\
Valouxis (3sc-) & 0.19 (11.3x) & 2.15 & 0 \\
Valouxis (sc+) & 0.50 (5.7x) & 2.83 & +2 \\
Valouxis (3sc+) & 0.61 (5.1x) & 3.12 & +5 \\
Valouxis (e+) & 0.68 (4.8x) & 3.23 & +3 \\
Valouxis (3e+) & 0.77 (4.6x) & 3.51 & +7 \\
Valouxis (e-) & 0.62 (6.1x) & 3.81 & +5 \\
Valouxis (3e-) & 0.71 (5.3x) & 3.79 & +11 \\
\hline\hline
Average & 0.54 (5.7x) & 3.1 & +4.1 \\
\hline
\end{tabular} 
\end{table} 

\begin{table}[!htb]
\catcode`\-=12
\centering
\caption{\label{tab:exp_ba2}Results of experiments with the complex classifiers: two-layer neural network}
\begin{tabular}{|l|r|r|r|r|r|}
\hline
\multirow{2}{*}{Instance}  & \multicolumn{3}{|c|}{CPU time [s]} & \multicolumn{2}{|c|}{Difference in solution quality} \\
\cline{2-6}
& WaldBoost & AdaBoost & standard & WaldBoost & AdaBoost \\
\hline\hline
Valouxis & 1.11 (3.5x) & 1.29 (3.0x) & 3.86 & 0 & -1 \\
Millar & 0.05 (2.2x) & 0.06 (1.8x) & 0.11 & 0 & -1 \\
Millar-s & 0.02 (3.5x) & 0.04 (1.8x) & 0.07 & 0 & -1 \\
Ortec & 3.24 (1.6x) & 3.89 (1.3x) & 5.23 & +1 & +1 \\
GPost & 0.61 (1.8x) & 0.97 (1.1x) & 1.07 & +2 & +1 \\
\hline\hline
Average & 1.00 (2.1x) & 1.25 (1.7x) & 2.07 & +0.6 & -0.2\\
\hline\hline
bp01 & 252 (2.9x) & 398 (1.8x) & 721 & +4 & +3 \\
bp02 & 214 (3.9x) & 298 (2.8x) & 833 & +7 & +5 \\
bp03 & 336 (3.3x) & 427 (2.6x) & 1110 & +11 & +8 \\
bp04 & 207 (4.3x) & 356 (2.5x) & 890 & +8 & +5 \\
bp05 & 224 (3.1x) & 365 (1.9x) & 693 & +5 & +4 \\
\hline\hline
Average & 246.6 (3.4x) & 368.8 (2.3x) & 849.4 & +7 & +5 \\
\hline
\end{tabular} 
\end{table}

\begin{table}[!htb]
\catcode`\-=12
\centering
\caption{\label{tab:exp_ba1}Results of experiments with complex classifiers: one-layer neural network}
\begin{tabular}{|l|r|r|r|r|r|}
\hline
\multirow{2}{*}{Instance}  & \multicolumn{3}{|c|}{CPU time [s]} & \multicolumn{2}{|c|}{Difference in solution quality} \\
\cline{2-6}
& WaldBoost & AdaBoost & standard & WaldBoost & AdaBoost \\
\hline\hline
Valouxis & 0.41 (9.4x) & 0.61 (6.3x) & 3.86 & +1 & +1 \\
Millar & 0.02 (5.5x) & 0.03 (3.7x) & 0.11 & +1 & 0 \\
Millar-s & 0.01 (7x) & 0.02 (3.5x) & 0.07 & 0 & 0 \\
Ortec & 2.07 (2.5x) & 2.79 (1.9x) & 5.23 & +4 & +3 \\
GPost & 0.32 (3.3x) & 0.53 (2.0x) & 1.07 & +5 & +3 \\
\hline\hline
Average & 0.56 (3.7x) & 0.80 (2.6x) & 2.07 & +2.2 & +1.4\\
\hline\hline
bp01 & 148 (4.9x) & 301 (3.4x) & 721 & +7 & +6 \\
bp02 & 122 (6.8x) & 176 (4.7x) & 833 & +10 & +9 \\
bp03 & 178 (6.2x) & 289 (3.8x) & 1110 & +14 & +12 \\
bp04 & 131 (6.8x) & 213 (4.2x) & 890 & +11 & +8 \\
bp05 & 109 (6.4x) & 201 (3.4x) & 693 & +8 & +8 \\
\hline\hline
Average & 137.6 (6.2x) & 236 (3.6x) & 849.4 & +10 & +8.6 \\
\hline
\end{tabular} 
\end{table}

The results obtained with the complex classifiers are shown in Table~\ref{tab:exp_ba2} and Table~\ref{tab:exp_ba1}, where two-layer and one-layer neural networks were used as the weak classifiers, respectively. Table~\ref{tab:exp_mcc} shows the misclassification rates for both types of neural networks. The contents of the columns are the same as those described for the previous tables. The weak classifiers used in the experiments are described in Section \ref{sec:wc}.

\begin{table}[!htb]
\catcode`\-=12
\centering
\caption{\label{tab:exp_mcc}Misclassification rates of the complex classifiers}
\begin{tabular}{|l|r|r|r|r|r|r|r|r|}
\hline
\multirow{4}{*}{Instance} & \multicolumn{8}{|c|}{Misclassification rate [\%]} \\
\cline{2-9}
& \multicolumn{4}{|c|}{1-layer} & \multicolumn{4}{|c|}{2-layer} \\
\cline{2-9}
& \multicolumn{2}{|c|}{AdaBoost} & \multicolumn{2}{|c|}{WaldBoost} & \multicolumn{2}{|c|}{AdaBoost} & \multicolumn{2}{|c|}{WaldBoost} \\
\cline{2-9}
& reject & accept & reject & accept & reject & accept & reject & accept \\
\cline{2-9}
\hline\hline
Valouxis & 7.1 & 15.6 & 7.3 & 16.1 & 6.5 & 15.1 & 6.9 & 15.4 \\
Millar & 6.3 & 13.9 & 6.7 & 14.3 & 6.0 & 13.5 & 6.1 & 13.8 \\
Millar-s & 6.1 & 13.2 & 6.3 & 13.9 & 5.7 & 12.6 & 5.8 & 13.3 \\
Ortec & 7.8 & 14.7 & 8.1 & 14.9 & 7.1 & 14.3 & 7.4 & 14.3 \\
GPost & 7.3 & 15.1 & 8.0 & 15.8 & 6.9 & 14.6 & 7.3 & 15.2 \\
\hline\hline
Average & 6.9 & 14.5 & 7.3 & 15.0 & 6.5 & 14.0 & 6.7 & 14.4 \\
\hline\hline
bp01 & 8.2 & 16.8 & 8.4 & 17.2 & 7.5 & 16.1 & 7.6 & 16.5 \\
bp02 & 7.7 & 14.6 & 8.0 & 15.3 & 7.1 & 14.2 & 7.2 & 14.4 \\
bp03 & 7.6 & 13.9 & 7.9 & 14.8 & 6.8 & 13.2 & 7.2 & 14.0 \\
bp04 & 8.1 & 15.5 & 8.3 & 16.3 & 7.2 & 14.6 & 7.6 & 15.5 \\
bp05 & 8.3 & 15.8 & 8.8 & 16.4 & 7.7 & 15.5 & 8.4 & 15.7 \\
\hline\hline
Average & 8.0 & 15.3 & 8.3 & 16.0 & 7.3 & 14.7 & 7.6 & 15.2 \\
\hline
\end{tabular} 
\end{table}

\subsection{Impact of the Classifier Evaluation on the Tabu Search Algorithm}
\label{sec:disc}
\begin{figure}[!htb]
\centering
\includegraphics[width=\linewidth]{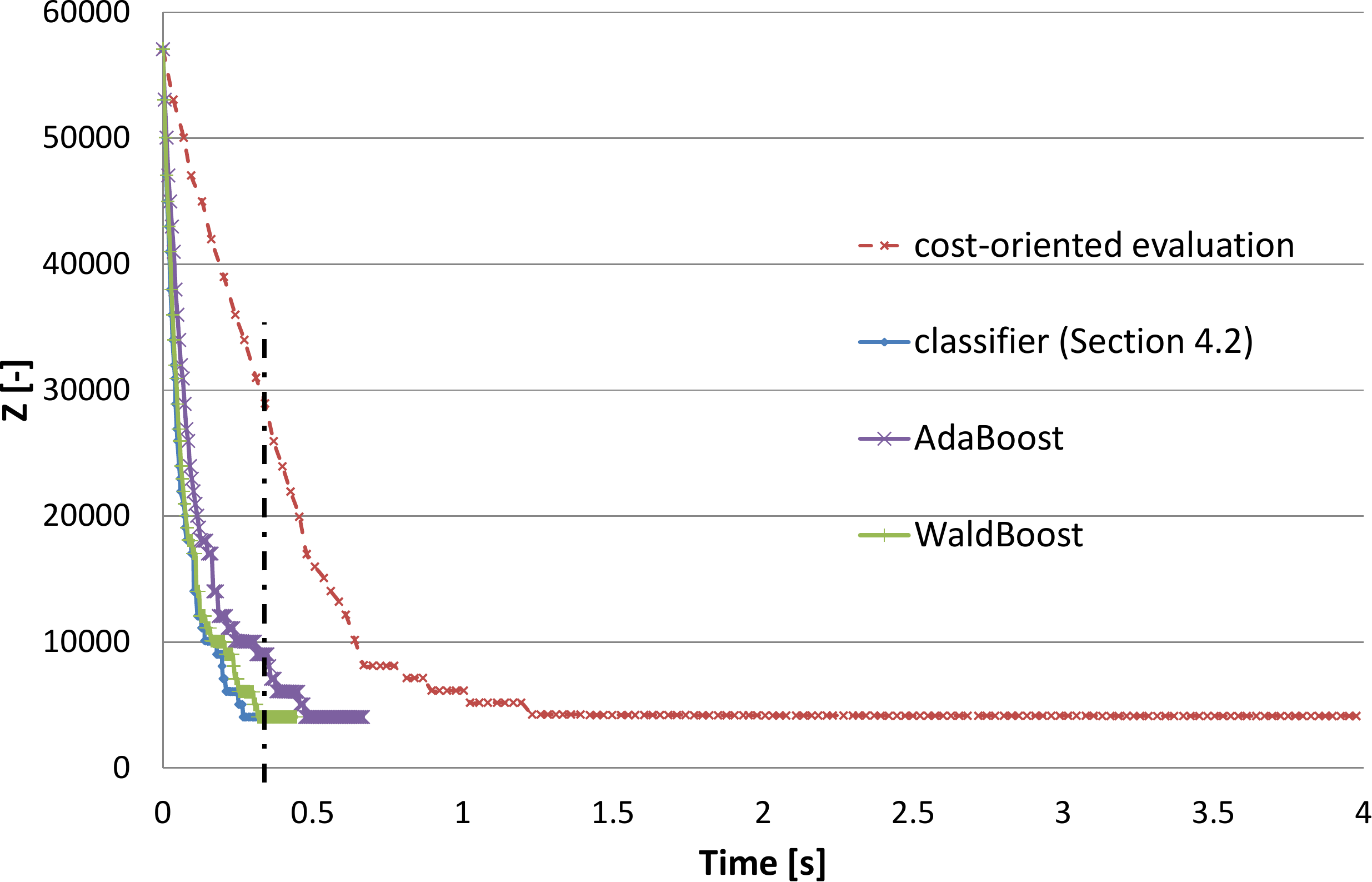}
\caption{Application of the tabu search algorithm to the Valouxis problem instance -- progress of the objective value over time for the cost-oriented evaluation and the evaluation using the classifiers}
\label{fig:graph}
\end{figure}

Figure~\ref{fig:graph} demonstrates the impact of our approach on the performance of the TSA, i.e., for the Valouxis instance (further results for different instances can be found in Appendix~I), which shows the progress of the objective function value during the TSA runtime. Clearly, speedup was achieved while the quality of the final solution was comparable. The simple classifier completed its run in 0.4 seconds and the value of the cost-oriented objective function was five times higher than the achieved (sub)optimal value from the simple classifier approach. It should be noted that the same stopping criterion was used in both cases. Furthermore, the graph shows the slow convergence of the cost-oriented evaluation compared with the classifiers.

Overall, the experiments confirmed that the classification rates were good:
\begin{itemize}
	\item standard instances -- $86.93\%$ for the simple classifiers, $89.85\%$ for AdaBoost, and $88.97\%$ for WaldBoost; 
	\item real-world instances -- $84.11\%$ for the simple classifiers, $86.97\%$ for AdaBoost, and $86.48\%$ for WaldBoost.
\end{itemize} 
The average speedup rates were roughly:
\begin{itemize}
	\item standard instances -- four times for the simple classifiers and two times for the complex classifiers with a two-layer neural network, where the solution quality remained similar in both cases;
	\item real-world instances -- seven times for the simple classifiers and three times for the complex classifiers with a two-layer neural network, where the solution quality was slightly worse in both cases.	
\end{itemize} 
The classifications rates were lower for the real-world instances because these instances were mostly over-constrained, and thus more sensitive to small changes in the pattern $p$. Moreover, these instances are much larger than the standard benchmark instances, e.g., in terms of the difference in the number of shift types (see Table~\ref{tab:desc}). The speedup was better for the real-world instances because:
\begin{enumerate}
	\item the constraints were more complex and time demanding, which led to a longer standard cost-oriented evaluation (note that the classifier is not dependent on the complexity and the number of constraints in terms of the time needed for classification); and
	\item the algorithm runtime was longer, so there was significant effect of employing the classifier.
\end{enumerate}

Furthermore, the misclassification rate for the incorrectly rejected solutions was better than that for the incorrectly accepted solutions because it was easier to discover bad solutions/structures based on the classifier.

\begin{figure}[!htb]
\centering\includegraphics[width=\linewidth]{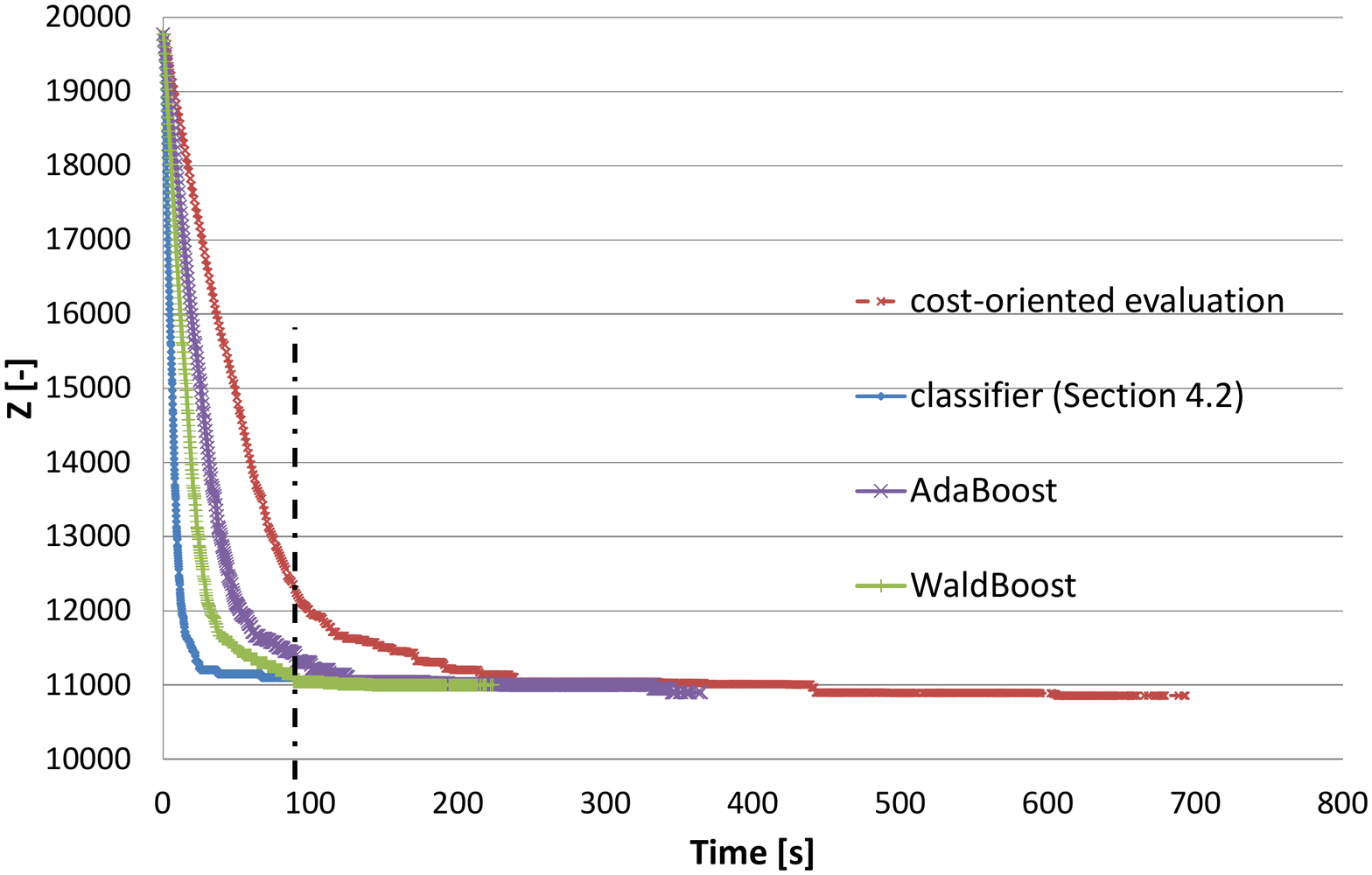}
\caption{bp01 problem instance -- progress of the objective value over time for the cost-oriented evaluation and the evaluations using the classifiers}
\label{fig:exp_bp01}
\end{figure}

\begin{figure}[!htb]
\centering
\includegraphics[width=\linewidth]{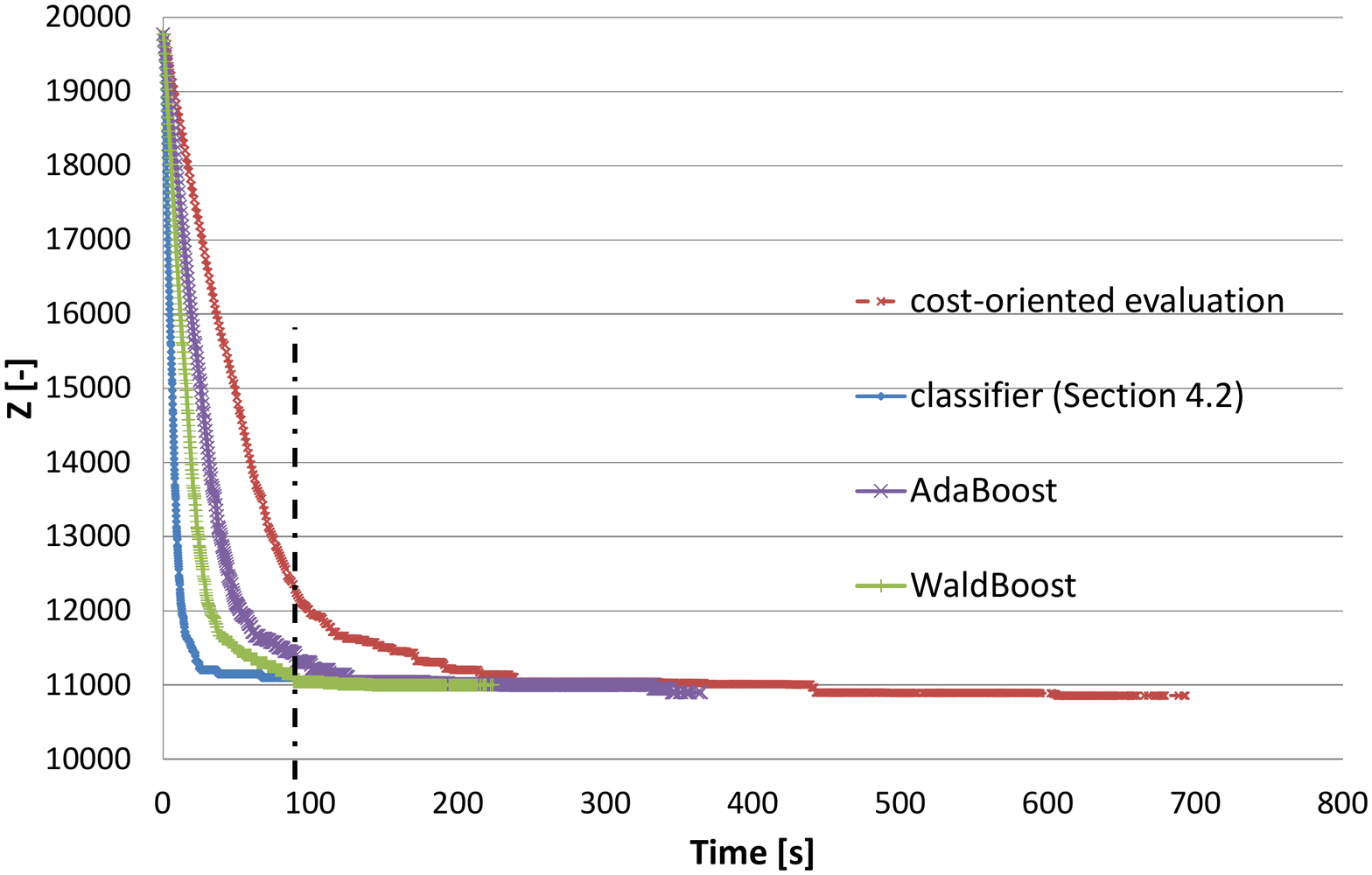}
\caption{pb05 problem instance -- progress of the objective value over time for the cost-oriented evaluation and the evaluations using the classifiers}
\label{fig:exp_bp05}
\end{figure}

Table~\ref{tab:exp} shows that the difference in the solution quality for the standard instances (i.e., small and medium instances) was not an issue because the positive differences (lower quality) were negligible. Indeed, the quality was even better in some cases, e.g., the Valouxis benchmark instance. By contrast, the difference in the solution quality was perceptible for the real-world instances (i.e., large instances). However, although the quality was slightly worse, the result was still completely acceptable with our approach in terms of the trade-off between time and quality because of the following reasons:
\begin{enumerate}
	\item the algorithm with the classifier obtained the solution much faster, i.e., the response time of the server was significantly better (see Figure~\ref{fig:motivation});
	\item the difference in quality was in the state where the solution was practically (sub)optimal, and thus the difference did not have a huge impact (see Figures~\ref{fig:exp_bp01}--\ref{fig:exp_bp05});
	\item the small difference in quality could not be recognized by a human being;
	\item the quality of the final roster was still very good and it could be used in practice.
\end{enumerate}

The boosting algorithms were slower than the simple classifiers (two or more times), but they were more accurate (approximately 3\%). Basically, there must be a compromise between a better quality roster and the longer processing time of the scheduling algorithm. WaldBoost was generally faster than AdaBoost due to the use of a sequential probability ratio test, but the roster quality obtained was slightly worse.

The robustness of our approach was analyzed based on the Valouxis instance (see Table~\ref{tab:exp_modif}). Clearly, the removal of the constraints had almost no impact on the solution quality. This was because the classifier still made stronger decisions than were currently required. However, adding constraints caused difficulties in some cases because the classifier was much weaker in terms of the evaluation (especially when the new constraint was very strict with a high weight). The addition or removal of employees (considering the same coverage of shifts) had a very similar effect, where the solution quality degraded increasingly as the number of added/removed employees increased. This was because the learned patterns became more irrelevant due to the complete change in the structure (the distribution of shifts among the available employees). However, when the coverage of shifts was adjusted according to the change in personnel, then the differences in the solution quality were negligible. The limit of possible changes in terms of adding/removing soft constraints/employees cannot be determined globally because every instance is different and it is not clear whether they will be highly sensitive to such changes.

Finally, two important observations can be made, based on the experimental results. The trained classifier can be successfully used: 1) on the same instances with data different from the training data; and 2) on similar instances with added/removed soft constraints/employees. Moreover, our solution does not place demands on the actual constraints: (i) various types of constraints can be used, as shown by the experiments where the real-world instances had completely different constraints than the standard instances; and (ii) the complexity of the constraints does not influence the time needed for the evaluation (see Table~\ref{tab:exp_ep}). Our approach can also handle various types of employee contracts (e.g., part-time vs. full-time, which leads to different soft constraints). This problem is solved by using several classifiers for each type of employee contract, e.g., the Ortec benchmark instance.

\subsection{Comparison and general assessment of the approach}
\label{sec:comp}
To the best of our knowledge, only one previous study \citep{Li12} has considered a learning approach that is comparable to our proposed method. However, a comparison of the results shows that there are some differences, as follows.
\begin{itemize}
	\item \cite{Li12} presented results (i.e., speedup and classification rate) only for one instance with different time periods, i.e., the Ortec benchmark instance. \textit{By contrast, we presented results (i.e., speedup, quality, and mis/classification rates) for various instances.}
	\item \cite{Li12} evaluated/classified the entire roster, \textit{whereas our method only considers the roster change for one nurse.}
	\item \cite{Li12} did not mention whether delta evaluation was used in the standard cost-oriented evaluation. \textit{We use ``$\Delta$ EC'' evaluation in the TSA.}
\end{itemize}

For the one-layer simple classifier (neural network), \cite{Li12} achieved a speedup of seven times and a classification rate of 83.2\%. \textit{With our method, the speedup was 15 times with a classification rate of 81.9\% for the Ortec instance and 19 times with an average classification rate of 83.0\% for various instances.} However, this classification rate was not sufficient to obtain a roster with reasonable quality, and thus the results obtained using the two-layer simple classifier and complex classifier are more relevant.

For the two-layer simple classifier (neural network), \cite{Li12} achieved a speedup of one time and a classification rate of 85.0\%. \textit{With our method, the speedup was three times with a classification rate of 85.22\% for the Ortec instance, and four times with a classification rate of 86.93\% for various instances.} The complex classifiers achieved a better classification rate (89.41\% on average) but the speedup was slightly worse (two times on average).

Our proposed approach could be applied in other domains of combinatorial optimization because the main idea (see Section \ref{sec:nnu}) is not bounded by the problem itself, but instead it is connected more closely to the specific structure of the problem solution. For example, exam timetabling \citep{Carter96} may be a good candidate because structural patterns can be found in the solutions and a time-demanding evaluation procedure is often used (after each change in the solution).

\section{Conclusions}
In this study, we showed that the use of learning techniques can be beneficial for solving personnel scheduling problems. We employ a neural network as the learning technique, where it works as a classifier. The neural network estimates whether a simple change in the roster of one employee is good (i.e., leading to an improvement in the objective function) or not. This approach was demonstrated using the TSA where it was used as a filter to eliminate most of the potentially bad candidate solutions. The remaining solutions were evaluated by the cost-oriented objective function. Our experiments demonstrated that algorithm runtime reductions were achieved (speedups of up to 10 times) while maintaining solutions with comparable quality (+/- a few percent). 

The major contribution of this study is the novel application of a classifier to determine a roster's quality (see Section \ref{sec:nnu}) and the practical integration of our approach in the TSA (see Section \ref{sec:cts}). The classification rates and speedups obtained using our method are better than the results reported by \cite{Li12}, mainly because the considered input pattern (the roster change for one employee versus the whole roster). In addition, we showed how our approach can be used in other (meta-, hyper-)heuristics, i.e., hill climbing and simulated annealing. The other contributions of our study include the original design of weak classifiers for boosting algorithms (see Section \ref{sec:wc}) and we provide an example of how the server-based rostering system can benefit from the proposed approach (see Section \ref{sec:m}).

Our main focus was to obtain good speedups for existing heuristics on NRPs. The properties of our approach were demonstrated extensively by the results of the experiments. In future research, these speedups can be further exploited in heuristics to facilitate better intensification and diversification processes. Furthermore, our approach can also be applied in other domains, such as exam timetabling, where the evaluation procedure for the intermediate solutions is time demanding and structural patterns can be identified in the solutions.

\begin{acknowledgements}
This work was supported by ARTEMIS FP7 EU and by the Ministry of Education of the Czech Republic under the project DEMANES 295372 and by the Grant Agency of the Czech Republic under the Project GACR FOREST P103-16-23509S. This is a post-peer-review, pre-copyedit version of an article published in Journal of Heuristics. The final authenticated version is available online at: \url{https://doi.org/10.1007/s10732-016-9314-9}.
\end{acknowledgements}

\clearpage
\section{Appendix I: Progress of the Tabu Search Algorithm on the Benchmark Instances}

\begin{figure}[!htb]
\centering
\includegraphics[width=10cm]{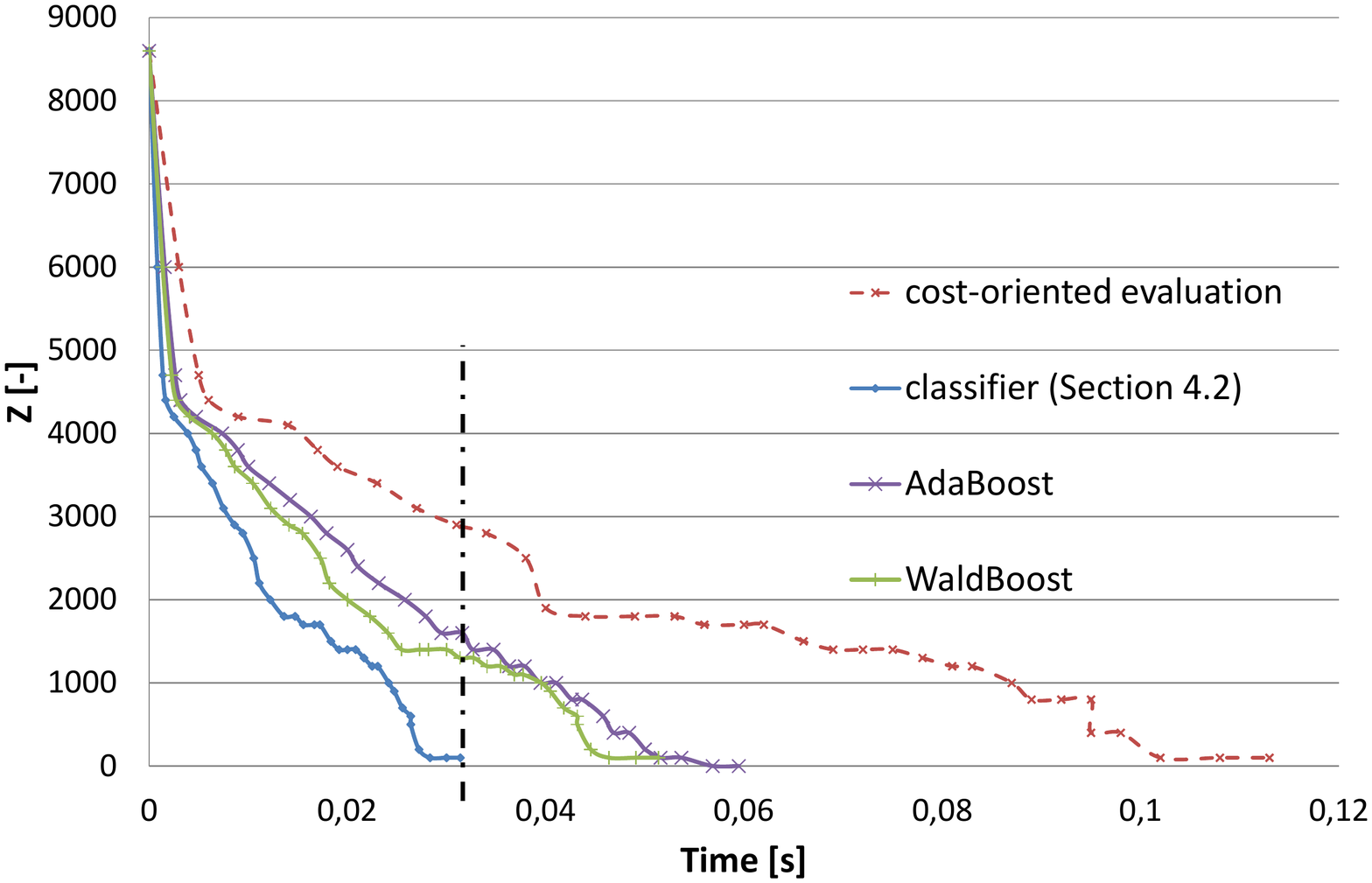}
\caption{Millar's problem instance -- progress of the objective value over time for the cost-oriented evaluation and the evaluations using the classifiers}
\end{figure}

\begin{figure}[!htb]
\centering\includegraphics[width=10cm]{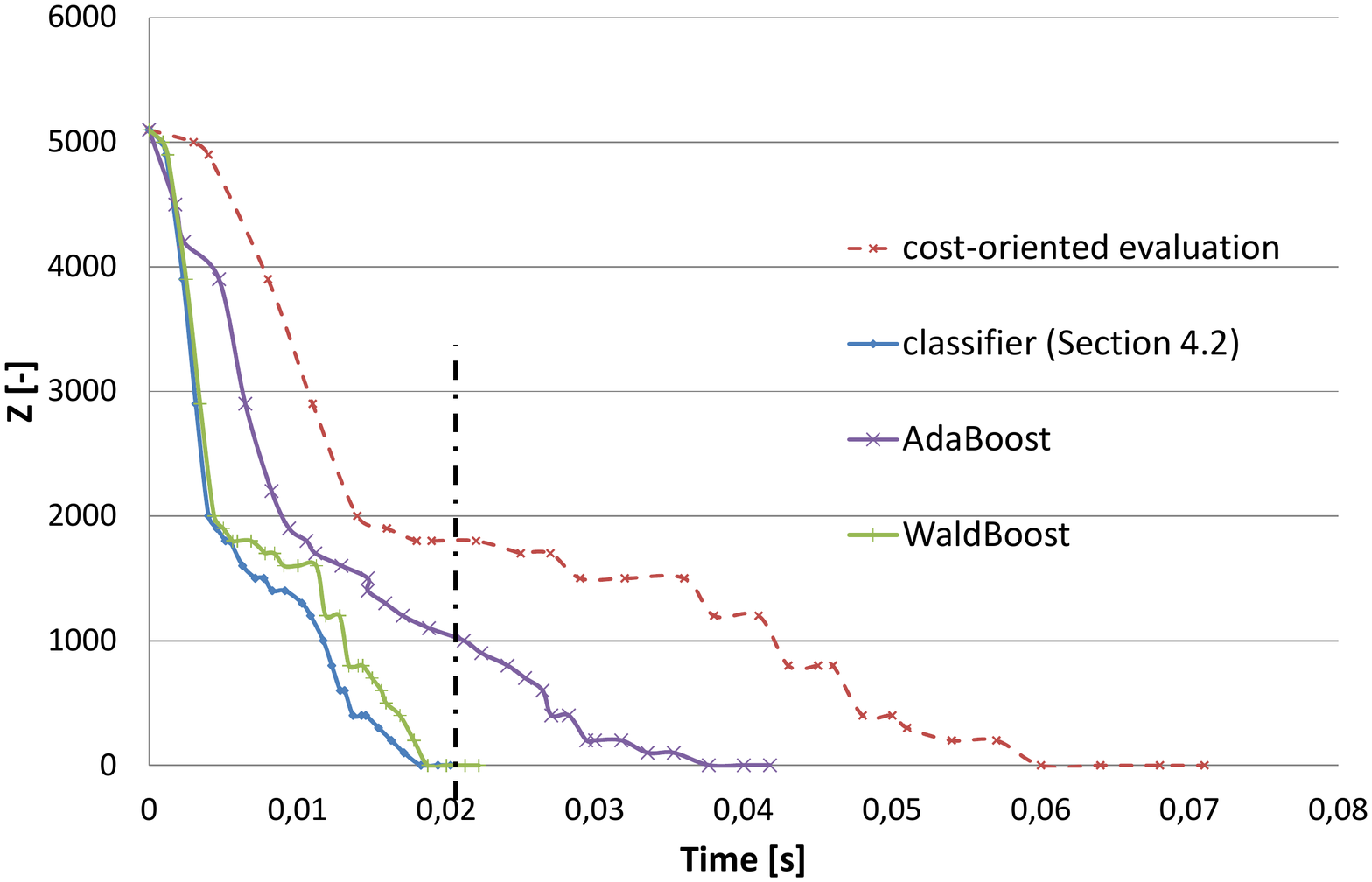}
\caption{Millar-s's problem instance -- progress of the objective value over time for the cost-oriented evaluation and the evaluations using the classifiers}
\end{figure}

\begin{figure}[!htb]
\centering\includegraphics[width=10.5cm]{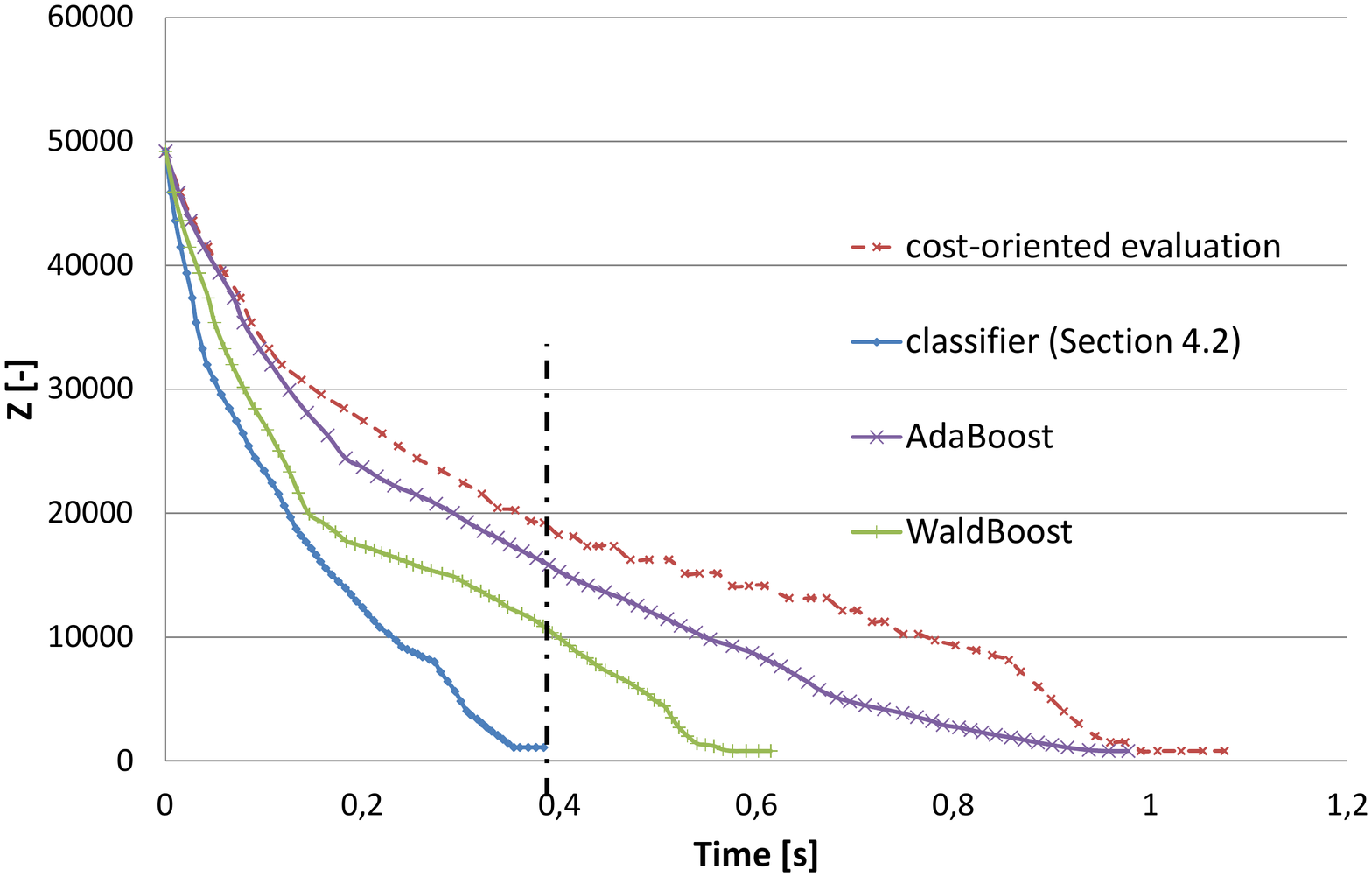}
\caption{Gpost problem instance -- progress of the objective value over time for the cost-oriented evaluation and the evaluations using the classifiers}
\end{figure}

\begin{figure}[!htb]
\centering\includegraphics[width=10.5cm]{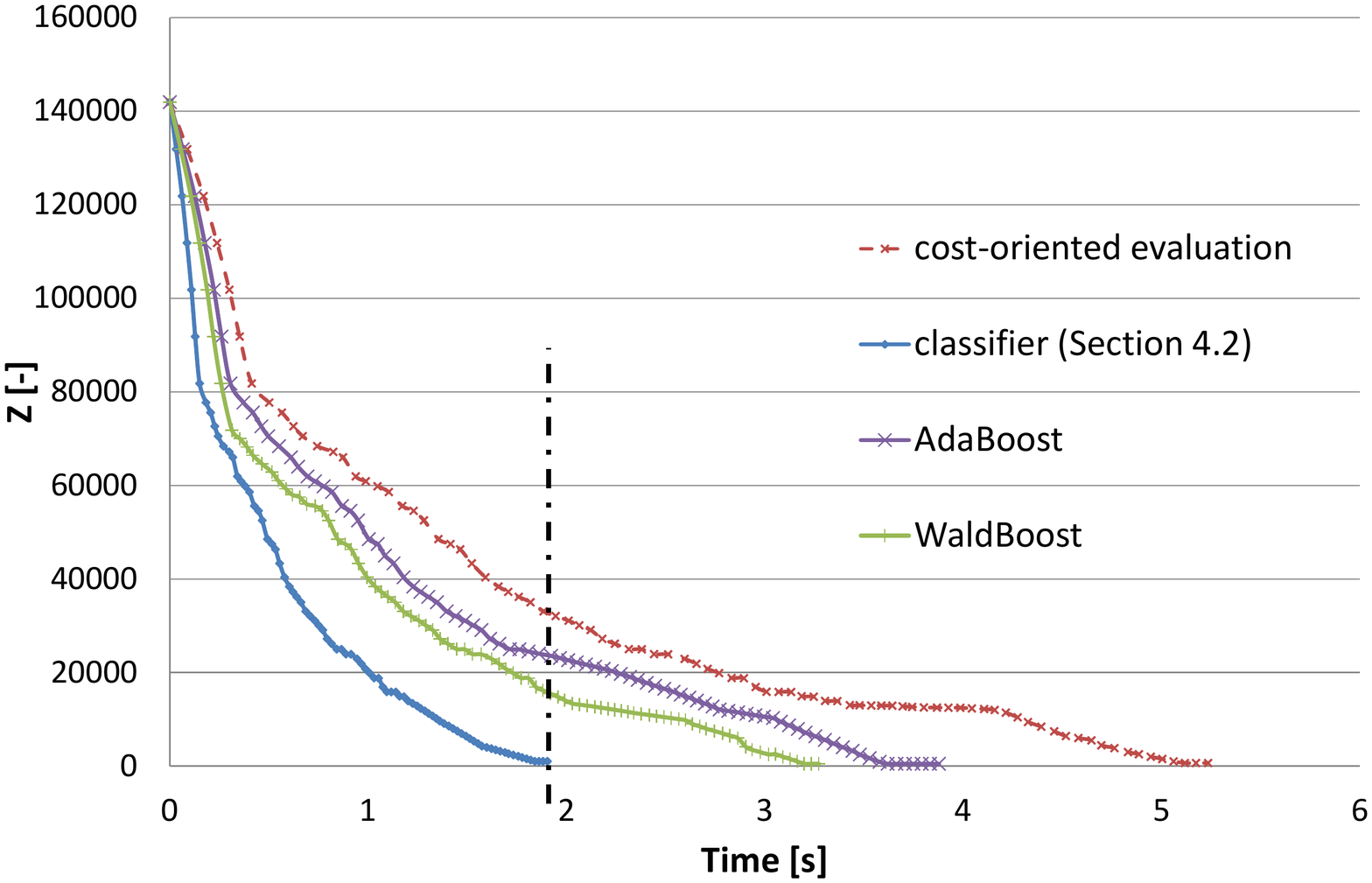}
\caption{Ortec problem instance -- progress of the objective value over time for the cost-oriented evaluation and the evaluations using the classifiers}
\end{figure}

\begin{figure}[!htb]
\centering
\includegraphics[width=\linewidth]{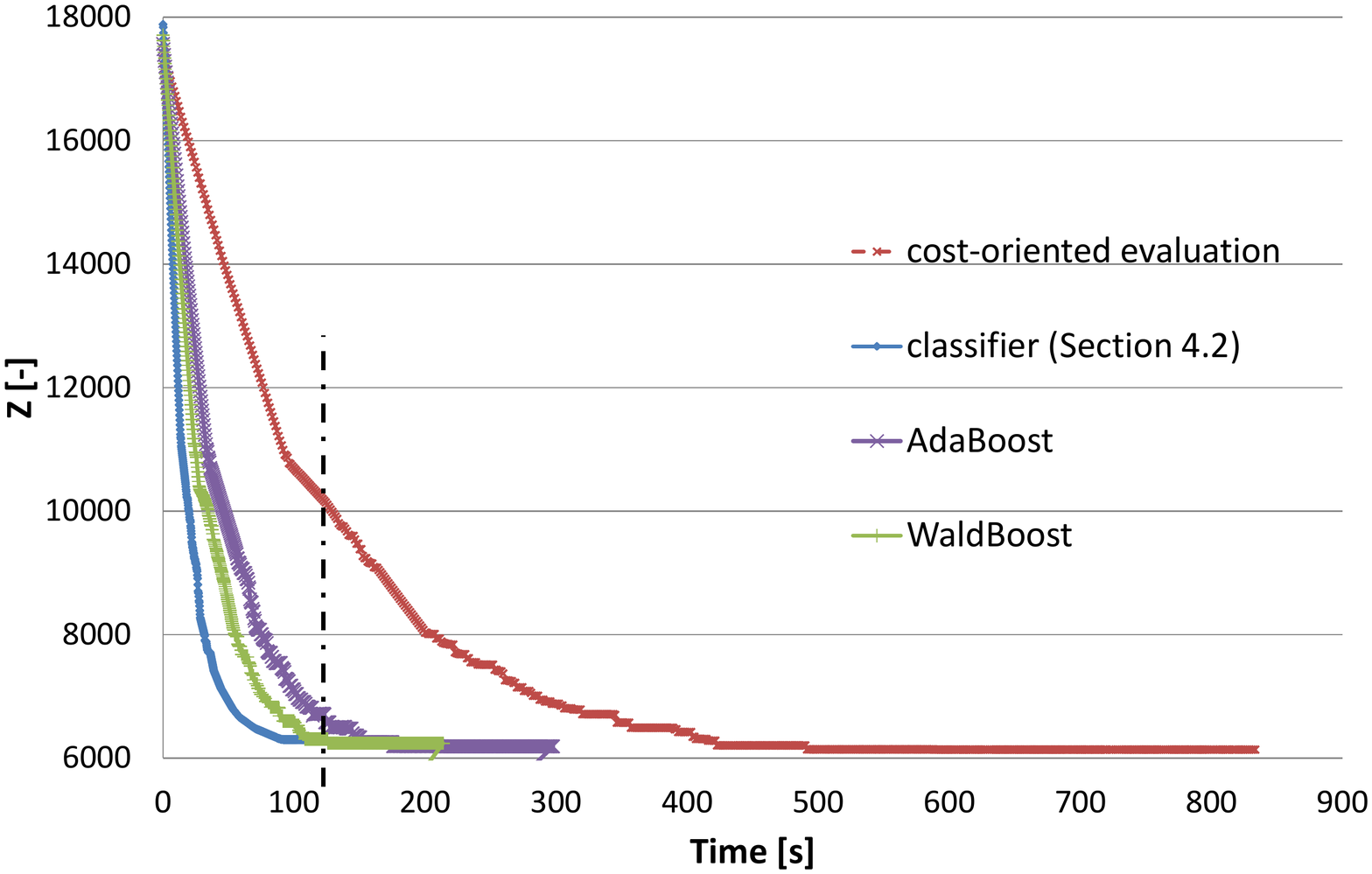}
\caption{bp02 problem instance -- progress of the objective value over time for the cost-oriented evaluation and the evaluations using the classifiers}
\end{figure}

\begin{figure}[!htb]
\centering
\includegraphics[width=\linewidth]{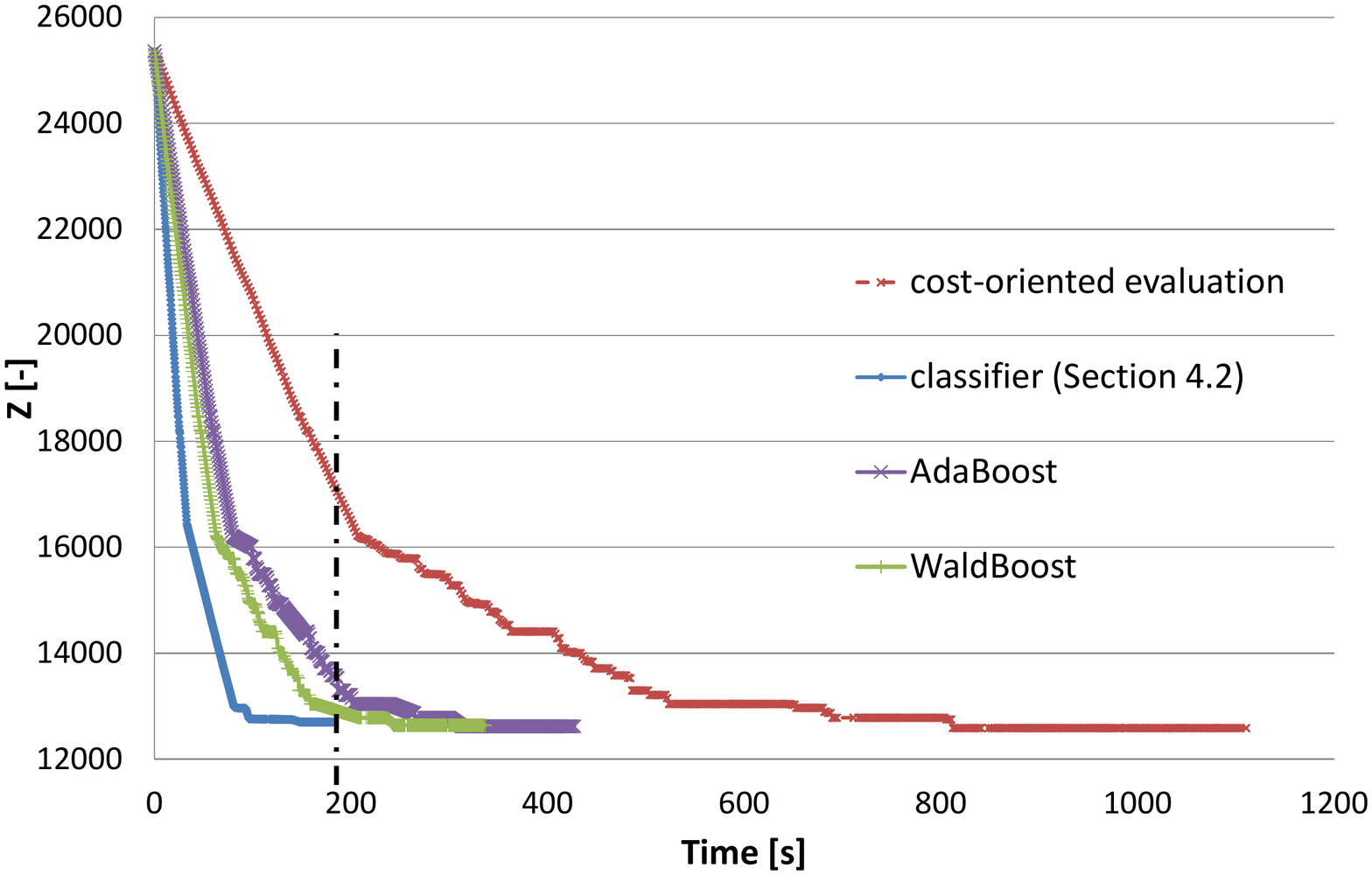}
\caption{bp03 problem instance -- progress of the objective value over time for the cost-oriented evaluation and the evaluations using the classifiers}
\end{figure}

\begin{figure}[!htb]
\centering
\includegraphics[width=\linewidth]{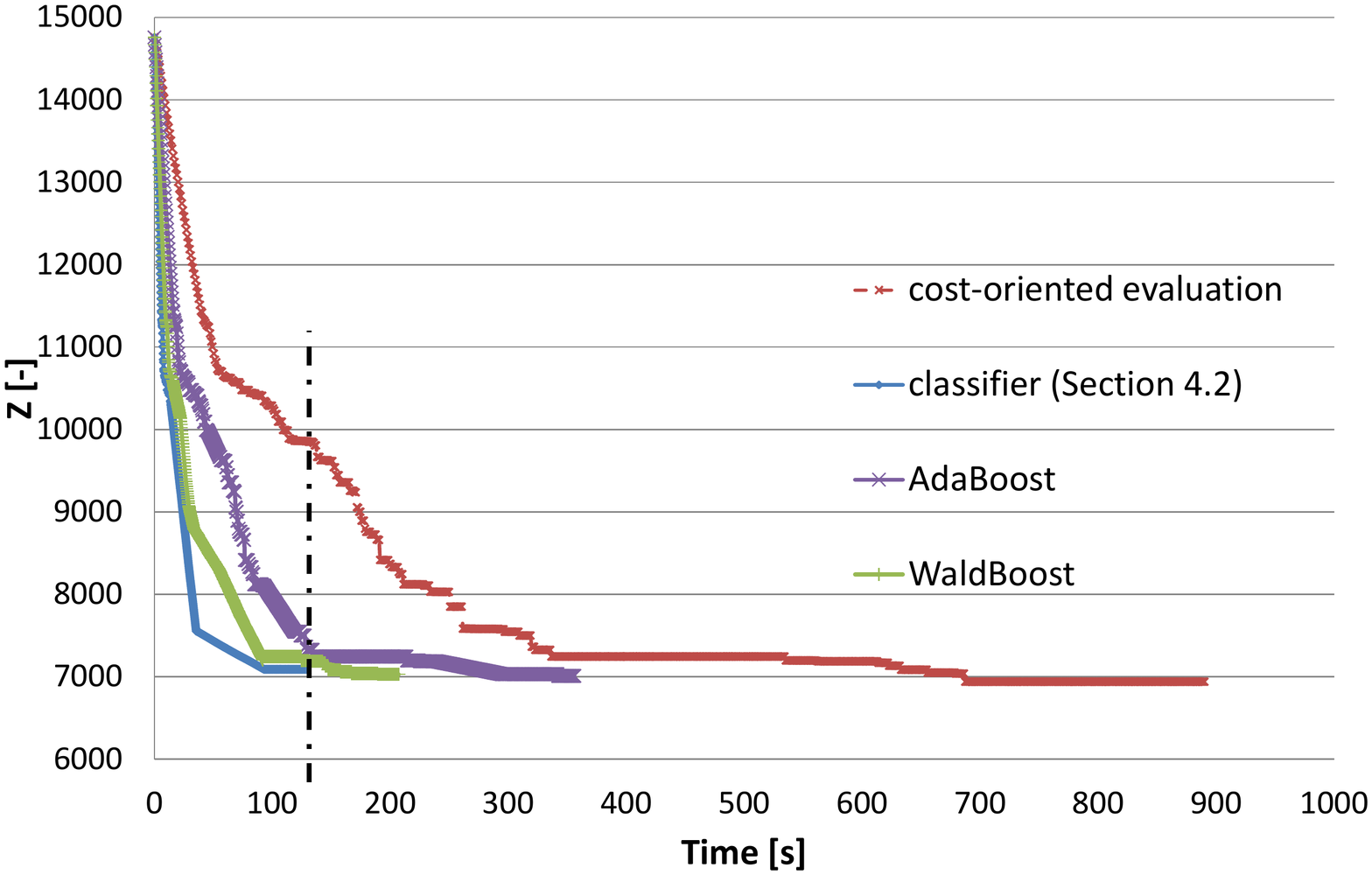}
\caption{pb04 problem instance -- progress of the objective value over time for the cost-oriented evaluation and the evaluations using the classifiers}
\end{figure}

\clearpage
\bibliographystyle{myspbasic}      
\bibliography{reference}   

\end{document}